\documentclass[sigconf]{acmart}
\pdfoutput=1
\AtBeginDocument{%
  \providecommand\BibTeX{{%
    \normalfont B\kern-0.5em{\scshape i\kern-0.25em b}\kern-0.8em\TeX}}}

\AtBeginDocument{%
\providecommand\BibTeX{{%
Bib\TeX}}}

\copyrightyear{2023}
\acmYear{2023}
\setcopyright{rightsretained}
\acmConference[CIKM '23]{Proceedings of the 32nd ACM International Conference on Information and Knowledge Management}{October 21--25, 2023}{Birmingham, United Kingdom}
\acmBooktitle{Proceedings of the 32nd ACM International Conference on Information and Knowledge Management (CIKM '23), October 21--25, 2023, Birmingham, United Kingdom}\acmDOI{10.1145/3583780.3615101}
\acmISBN{979-8-4007-0124-5/23/10}

\usepackage{algpseudocode}
\usepackage[ruled,vlined,linesnumbered]{algorithm2e}
\usepackage{amsthm}
\usepackage{amsfonts}
\usepackage{amsmath}

\usepackage{subcaption}
\usepackage{balance}
\usepackage{booktabs}
\usepackage{xspace}
\usepackage{multirow}
\usepackage{textcomp}
\usepackage{siunitx}
\usepackage{xcolor}
\usepackage{enumitem}
\usepackage[T1]{fontenc}

\begin{document}

\title{Unleashing the Power of Shared Label Structures \\ for Human Activity Recognition}

\author{Xiyuan Zhang}
\affiliation{
  \institution{University of California, San Diego}
  \country{}
  }
\email{xiyuanzh@ucsd.edu}

\author{Ranak Roy Chowdhury}
\affiliation{
  \institution{University of California, San Diego}
  \country{}
  }
\email{rrchowdh@eng.ucsd.edu}

\author{Jiayun Zhang}
\affiliation{
  \institution{University of California, San Diego}
  \country{}
  }
\email{jiz069@ucsd.edu}

\author{Dezhi Hong}
\authornote{Work unrelated to Amazon.}
\affiliation{
  \institution{Amazon}
  \country{}
  }
\email{hondezhi@amazon.com}

\author{Rajesh K. Gupta}
\affiliation{
  \institution{University of California, San Diego}
  \country{}
  }
\email{rgupta@ucsd.edu}

\author{Jingbo Shang}
\affiliation{
  \institution{University of California, San Diego}
  \country{}
  }
\email{jshang@ucsd.edu}

\newcommand{\our}{\mbox{SHARE}\xspace}
\newcommand{\van}{\mbox{VanillaHAR}\xspace}
\newcommand{\smallsection}[1]{\vspace{1mm}\noindent\textbf{#1.}}
\newtheorem{assumption}{Assumption}

\newcommand{\delete}[1]{}

\renewcommand{\shorttitle}{Unleashing the Power of Shared Label Structures for Human Activity Recognition}

\renewcommand{\shortauthors}{Xiyuan Zhang et al.}

\begin{abstract}
    Current human activity recognition (HAR) techniques regard activity labels as integer class IDs without explicitly modeling the semantics of class labels. We observe that different activity names often have shared structures. For example, ``open door'' and ``open fridge'' both have ``open'' as the action; ``kicking soccer ball'' and ``playing tennis ball'' both have ``ball'' as the object. Such shared structures in label names can be translated to the similarity in sensory data and modeling common structures would help uncover knowledge across different activities, especially for activities with limited samples. In this paper, we propose \our, a HAR framework that takes into account shared structures of label names for different activities. To exploit the shared structures, \our comprises an encoder for extracting features from input sensory time series and a decoder for generating label names as a token sequence.
We also propose three label augmentation techniques to help the model more effectively capture semantic structures across activities, including a basic token-level augmentation, and two enhanced embedding-level and sequence-level augmentations utilizing the capabilities of pre-trained models. \our outperforms state-of-the-art HAR models in extensive experiments on seven HAR benchmark datasets.
We also evaluate in few-shot learning and label imbalance settings and observe even more significant performance gap.

\end{abstract}




\maketitle

\begin{figure}[t]
      \centering
          \includegraphics[width=0.48\textwidth]{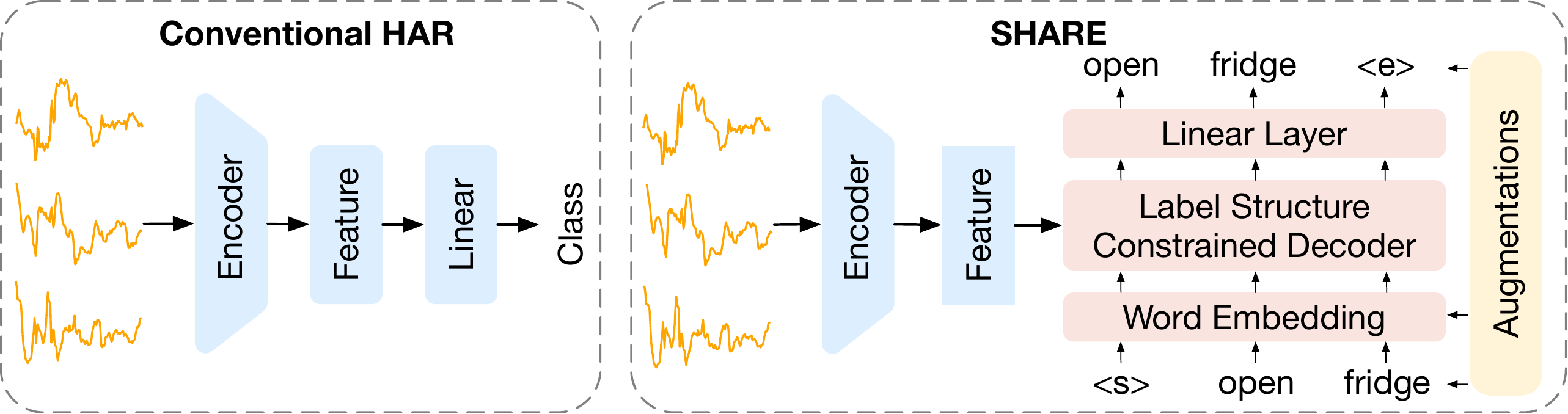}
         \caption{Existing HAR framework vs \our. \our exploits shared structures in label names and generates activity name sequences as prediction, rather than predicting integer class IDs. We also design three label augmentations at different levels to better capture shared structures.} 
    \label{fig:main}
    \vspace{-2mm}
\end{figure}

\section{Introduction}

\begin{figure*}[t]
      \centering
      \begin{subfigure}[c]{0.25\textwidth}
          \centering
          \includegraphics[width=\textwidth]{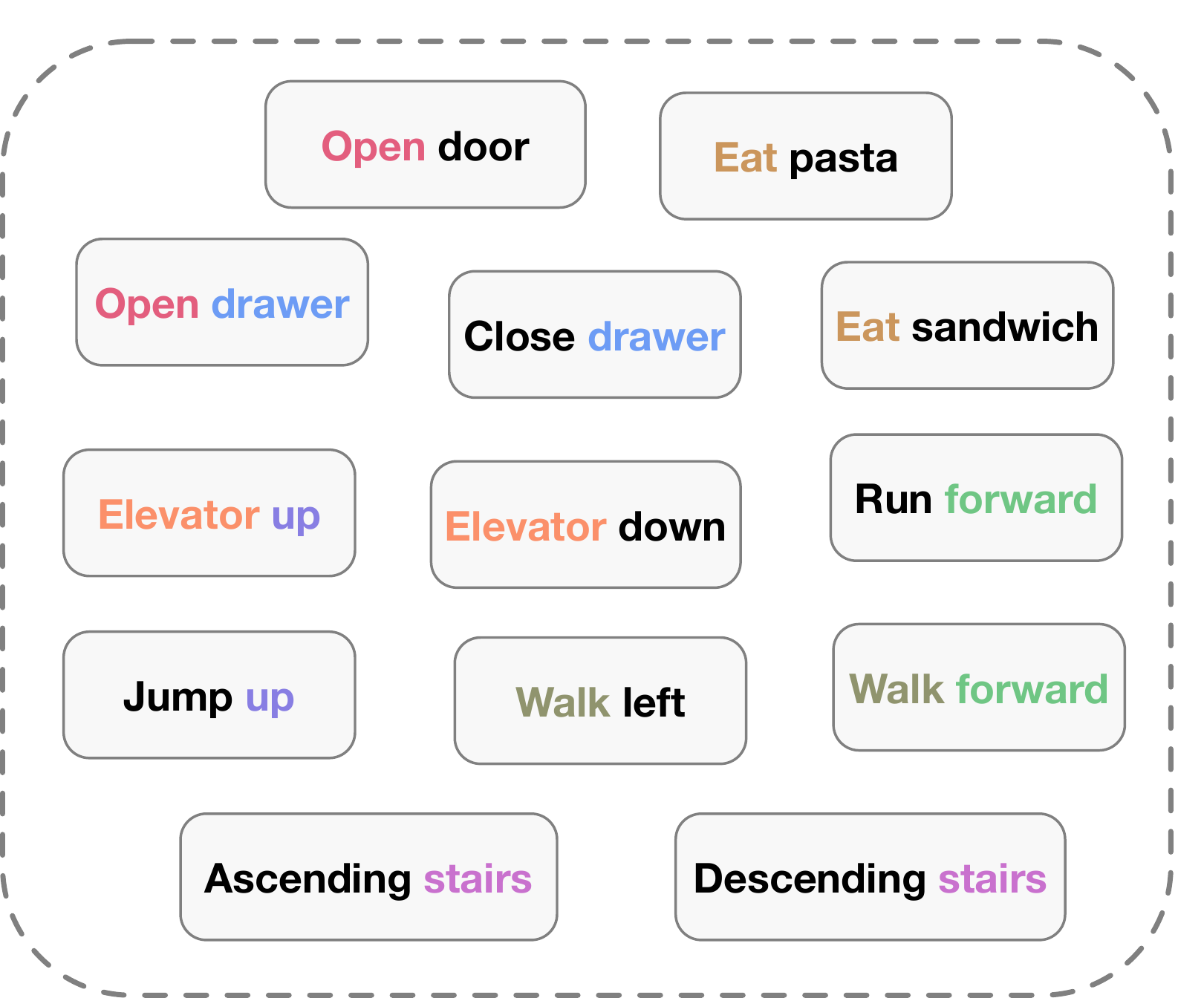}
          \caption{Label}
          \label{fig:label}
      \end{subfigure}
      \begin{subfigure}[c]{0.735\textwidth}
          \centering
          \includegraphics[width=\textwidth]{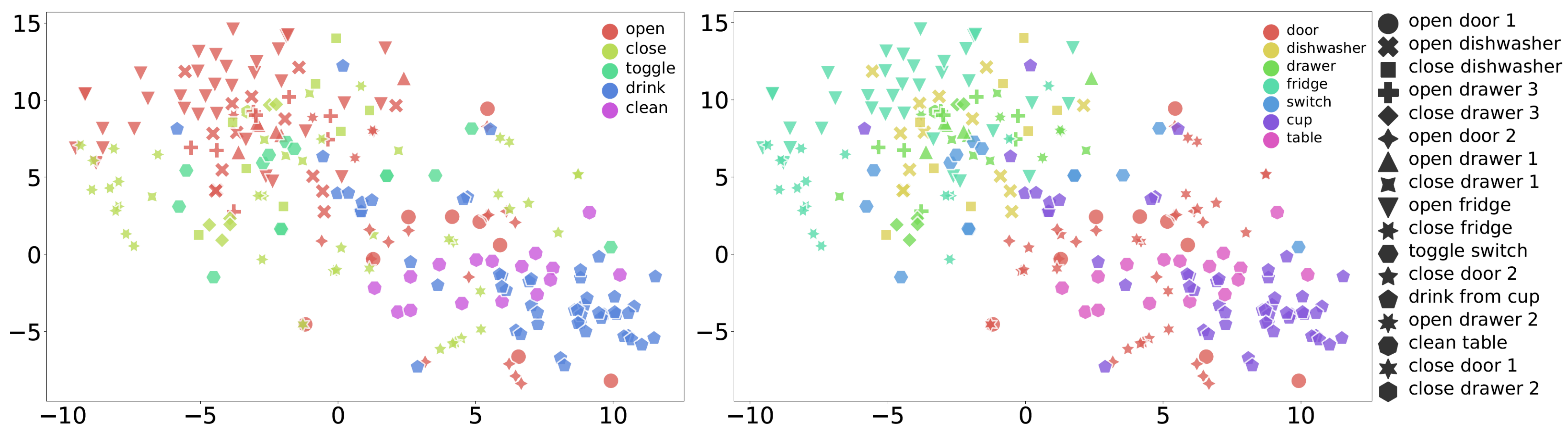}
          \caption{T-SNE results of sensory data with the colors denoting different shared label structures.}
          \label{fig:tsne}
      \end{subfigure}
         \caption{(a) Labels in HAR datasets typically share common structures. (b) T-SNE visualization of sensory data in Opportunity dataset~\cite{roggen2010collecting}. Activities with the same actions or objects (marked by the same colors) are closer. 
         Each point represents one data sample, and each type of marker represents a different type of activity. The two figures have the same set of data points and markers and only differ in colors. The same color represents common actions (left figure) or common objects (right figure). }
         \label{fig:intro}
\end{figure*}

Sensor-based human activity recognition (HAR) identifies human activities using sensor readings from wearable devices. HAR has a variety of applications including healthcare, motion tracking, smart home automation, human-computer interaction~\cite{hossain2019active,chikhaoui2017towards,chen2019distributionally,ma2019labelforest,li2021meta,yao2017deepsense}. 
For example, acceleration sensors attached to legs record subjects walking around and performing daily activities for gait analysis for Parkinson's disease patients~\cite{bachlin2009wearable}; accelerometer and gyroscope can monitor user postures to detect falls for elderly people~\cite{wibisono2013falls}. 

While tremendously valuable, HAR data remain difficult to collect due to security or privacy concerns, as human subjects involved in the collection process may not consent to data sharing or data transmission over the network. This often leads to local training at the edge using limited samples from just a few human subjects. Additionally, certain types of human activities happen less frequently by nature, further complicating data collection. 

We note that existing HAR methods treat labels simply as integer class IDs and learn their semantics purely from annotated sensor data. This is less effective especially when labeled data are limited. To achieve better recognition performance, prior research mostly is concentrated on designing better 
feature extraction modules~\cite{ordonez2016deep,fazli2021hhar,radu2018multimodal,li2021two} while largely overlooking the advantages of modeling label structures.
Since sensory readings measuring human activities are time-series data, existing time-series classification models are also applicable to HAR. These methods, however, 
are also primarily focused on enhancing feature extraction~\cite{dempster2020rocket,zerveas2021transformer,chowdhury2023primenet}.  It is noteworthy that both HAR and time-series classification methods in the literature miss the modeling of label name structures. 

We argue that a more effective approach to learning activity semantics is through label name modeling, as activity names in HAR datasets often share structures that reflect the similarity between different activities.
For example, both ``open door'' and ``open fridge'' (sharing the action ``open'') involve pulling a (fridge) door around a hinge (while ``open door'' first rotates the knob to release the lock and ``open fridge'' directly pulls the handle); both ``stairs up'' and ``stairs down'' (sharing the object ``stairs'') need to bend the knees and extend the legs.
Figure~\ref{fig:label} illustrates more examples of activity label names in typical HAR datasets (e.g., ``eat pasta'' and ``eat sandwich'', ``elevator up'' and  ``elevator down''). 
The common actions or objects in these examples translate to similarities in the IMU data space.
As shown in Figure~\ref{fig:tsne}, we apply t-SNE visualization on sensor readings from the Opportunity dataset~\cite{roggen2010collecting}. We color different activities by common actions or objects. 
Activities of the same color (sharing the same action or object in label names) appear closer in the embedding space, indicating stronger similarity in the original sensory measurements. 
Such mapping between input features and label names motivates us to design a more effective learning framework that extracts knowledge from label structures.

To this end, we propose \our, shown in Figure~\ref{fig:main}, which models both input sensory data features and label name structures. \our comprises an encoder for extracting features from sensory input and a decoder for predicting label names. 
Unlike existing HAR models that output integer class IDs as prediction results, \our outputs \emph{label name sequences}, thus preserving structures among various activities and providing a global view of activity relationships.
During training, we optimize the model by minimizing the differences between predicted label names and ground-truth label names. During inference, we exploit a constrained decoding method to produce only valid labels.

We also design three label augmentation methods at different levels to better capture shared structures across activities. The basic token-level augmentation randomly replaces the original label sequences by their meaningful tokens (e.g., all actions of ``eat X'' are treated as a class of ``eat''). This happens only during training and helps the model consolidate semantics of shared structures across different activities. 
We further develop two embedding- and sequence-level augmentations leveraging pre-trained models. At the embedding level, we integrate pre-trained word embeddings to capture shared semantic meanings not obvious in label names (e.g., the similarity between ``walk'' and ``run''). At the label sequence level, 
for HAR datasets that do not have shared structures in their original labels, we offer an automated label generation method to generate new labels with shared tokens while preserving the same semantic meanings, leveraging large language models. Specifically, we use OpenAI's GPT-4~\cite{OpenAI2023GPT4TR} to extend atomic, non-overlapping label names into sequences of meaningful tokens.
To the best of our knowledge, \our is the first solution to HAR classification via decoding label sequences. We evaluate \our on seven HAR benchmark datasets and observe the new state-of-the-art performance. We summarize our main contributions as follows:
\begin{itemize}[nosep,leftmargin=*]
    \item We find shared structures in label names map to similarity in the input data, leading to a more effective HAR framework, \our, by modeling label structures. \our captures knowledge across activities by uncovering information from label structures. 
    \item We propose three label augmentation methods, each targeting at a different level, to more effectively identify shared structures across activities. These include a basic token-level augmentation and two pre-trained model-enhanced augmentations at the embedding level and at the label sequence level.
    \item We evaluate \our on seven HAR benchmark datasets and observe the new state-of-the-art performance. We also conduct experiments under few-shot settings and label imbalance settings and observe even more significant performance improvement.
\end{itemize}

\vspace{-1mm}
\section{Related Work}

\subsection{Human Activity Recognition}

Existing HAR approaches can be categorized into statistical methods and deep learning-based methods~\cite{chen2021deep,zhang2021deep}.
Traditional methods are based on data dimensionality reduction, spectral feature transformation (e.g., Fourier transformation), kernel embeddings~\cite{qian2018sensor}, first-order logic~\cite{zeng2010knowledge} or handcrafted statistical feature extraction (e.g., mean, variance, maximum, minimum)~\cite{figo2010preprocessing}. 
These features are then used as input to shallow machine learning methods like SVM, and Random Forest.
In recent years, deep learning methods have advanced automatic feature extraction and have begun to substitute hand-crafted feature engineering in HAR~\cite{hammerla2016deep,fazli2021hhar,yao2017deepsense}
, including convolutional neural networks, recurrent neural network, attention mechanism, and their combinations.
DeepConvLSTM~\cite{ordonez2016deep} is composed of convolutional layers for feature extractors and recurrent layers for capturing temporal dynamics of the feature representations.
MA-CNN~\cite{radu2018multimodal} designs modality-specific architecture to first learn sensor-specific information and then unify different representations for activity recognition.  
SenseHAR~\cite{jeyakumar2019sensehar} proposes a sensor fusion model that maps raw sensory readings to a virtual activity sensor, which is a shared low-dimensional latent space.
AttnSense~\cite{ma2019attnsense} further integrates attention mechanism to convolutional neural network and gated recurrent units network. 
THAT~\cite{li2021two} proposes a two-stream convolution augmented Transformer model for capturing range-based patterns.
We shall note that these models focus on designing more effective feature extractors for better performance but neglect the semantic information in label names, which is the focus of this work.

\subsection{Time-Series Classification}
HAR data are time-stamped sensory series, enabling the use of time-series classification methods.
Existing time-series classification models fall into two categories: statistical methods and deep learning methods. Statistical methods are based on nearest neighbor~\cite{bagnall2018uea,shokoohi2015non}, dictionary classifier~\cite{schafer2017multivariate}, ensemble classifier~\cite{lines2018time,shifaz2020ts}, etc. These statistical methods are more robust to data scarcity but do not scale well when the feature numbers in high-dimensional space become huge. On the other hand, deep learning methods can extract features from high-dimensional data but require abundant data points to train an effective model. 

Convolutional Networks (FCN and ResNet)~\cite{wang2017time,ismail2020inceptiontime} and Recurrent Neural Networks~\cite{karim2017lstm,karim2019multivariate} show better performance compared with statistical methods.
TapNet~\cite{zhang2020tapnet} is an attentional prototype network that calculates the distance to class prototypes to learn feature representations. 
ShapeNet~\cite{li2021shapenet} performs shapelet selection by embedding shapelet candidates into a unified space and trains the network with cluster-wise triplet loss. 
SimTSC~\cite{zha2022towards} formulates time-series classification as a graph node classification problem and uses a graph neural network to model similarity information.
Recently, Rocket~\cite{dempster2020rocket} applies plenty of random convolution kernels for data transformation and attains state-of-the-art accuracy. MiniRocket~\cite{dempster2021minirocket} maintains the accuracy and improves the processing time of Rocket. 
TST~\cite{zerveas2021transformer} and TARNet~\cite{chowdhury2022tarnet} incorporate unsupervised representation learning which offers benefits over fully supervised methods on the downstream classification tasks. 
Similar to existing HAR methods, time-series classification models focus on designing more advanced feature extraction or unsupervised representation learning methods without taking into account the label semantics, whereas \our models the shared structures in the label set for more effective representation learning.

\subsection{Label Semantics Modeling}
Given label name semantics as prior knowledge, classification tasks could benefit from modeling such semantics through knowledge graph~\cite{von2019informed} or textual information~\cite{lei2015predicting,radford2021learning,zhou2021deep,Zhang_2023}. 
Tong et al.~\cite{tong2021zero} exploit knowledge from video action recognition models to construct an informative semantic space that relates seen and unseen activity classes. Recent works designed specifically for zero-shot learning in human activity recognition also combine semantic embeddings~\cite{matsuki2019characterizing,wu2020multi,wang2017zero}. 
However, these works mostly calculate the mean embeddings for labels with multiple words, which misses label structures and is suboptimal.
Unlike these works, \our preserves label structures and enables knowledge sharing through decoding label names for generic HAR.
\section{Preliminary}\label{sec:setting}

\begin{figure*}[t]
      \centering
          \includegraphics[width=0.9\textwidth]{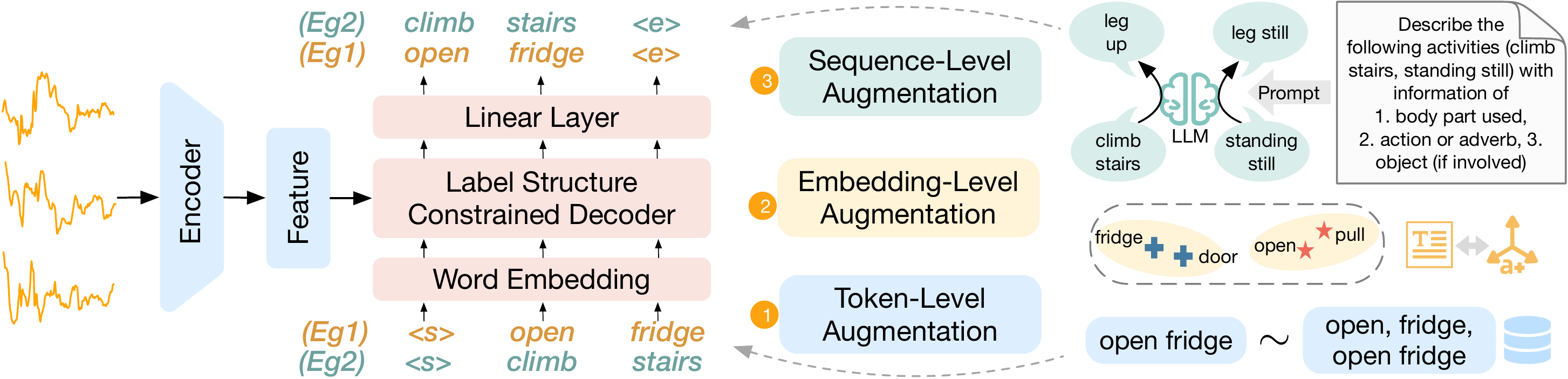}
         \caption{Framework of \our. We encode the time-series features and decode the label sequences as predictions. We further design three augmentation methods at different levels to better capture the shared semantic structures.} 
    \label{fig:model}
\end{figure*}

We focus on human activity recognition such as walking and sitting, captured by the sensory time-series data in a given time period. We formulate HAR settings of conventional methods and \our. 

\noindent \textbf{Conventional HAR.} We denote HAR dataset in conventional methods as $\mathcal{D'}=\{(\mathbf{x_i}, c_i)\}_{i=0}^N, \mathbf{x_i} \sim \mathcal{X}, c_i \sim \mathcal{C}$, where $\mathcal{X}$ and $\mathcal{C}$ denote the input space and the label space. Each sample of time-series input is denoted as $\mathbf{x_i} \in \mathbb{R}^{T_i \times v}$, where $T_i$ is the length of the time series, and $v$ is the number of measured variables. The label space $\mathcal{C}$ contains $C$ classes, and each label $c$ is an integer from $\{1,2,\cdots,C\}$.

\noindent \textbf{\our.} We denote dataset in \our as $\mathcal{D}=\{(\mathbf{x_i}, \mathbf{y_i})\}_{i=0}^N, \mathbf{x_i} \sim \mathcal{X}, \mathbf{y_i} \sim \mathcal{Y}$, where data space $\mathcal{X}$ is the same as conventional HAR methods, and $\mathcal{Y}$ denotes the label space in \our. We denote $\mathbf{y_i}=[y_{i1}, y_{i2}, \cdots, y_{ik_i}]$ as a sample human activity label sequence, where $k_i$ is the length of the label sequence $\mathbf{y_i}$. For example, the label ``walk upstairs'' contains a word sequence of length two, [``walk'', ``upstairs''] respectively. 
The label space $\mathcal{Y}$ contains $C$ classes and $M$ tokens. 
Instead of presenting labels as independent integer IDs, there exist shared structures across different labels in the label space $\mathcal{Y}$. For example, ``walk upstairs'' and ``walk downstairs'' both have ``walk''  in label names. Formally, there exist labels $\mathbf{y_i}, \mathbf{y_j}$, $i \neq j$ that have the same word $y_{im} = y_{jl}$, where $1 \leq m \leq k_i, 1 \leq l \leq k_j$ are positions in $\mathbf{y_i}$ and $\mathbf{y_j}$.

\begin{algorithm}[tb]
    \SetKwData{Left}{left}\SetKwData{This}{this}\SetKwData{Up}{up}
    \SetKwFunction{Union}{Union}\SetKwFunction{FindCompress}{FindCompress}
    \SetKwInOut{Input}{Input}\SetKwInOut{Output}{Output}\SetKwInOut{Model}{Model}
    \Input{Training set $\mathcal{D}_{\mathrm{tr}} = \{\mathbf{x_{i}}, \mathbf{y_{i}}\}_{i=0}^{N_{
\mathrm{tr}}}$, test set $\mathcal{D}_{\mathrm{te}} = \{\mathbf{x_{i}}, \mathbf{y_{i}}\}_{i=0}^{N_{\mathrm{te}}}$.}
\Model{Encoder $f_\theta$, Decoder $g_\phi$.} 
    \Output{Predicted label sequences on test set $\{\mathbf{\hat{y}_{i}}\}_{i=0}^{N_{\mathrm{te}}}$.}

    \If{no shared tokens in $\mathcal{Y}$}
    {Sequence-Level augmentation to $\mathcal{Y}$ \tcp*{Sec~\ref{sec:pretrain-aug}}}
    
    \While{not converge}{
    Sample $(\mathbf{x_i, y_i}) \sim \mathcal{D}_{\mathrm{tr}}$; 
    
    Token-Level augmentation $\mathbf{y'_{i}} \leftarrow \mathbf{y_{i}}$ \tcp*{Sec~\ref{sec:token-aug}}
    Encoder feature extraction $\mathbf{z_i}=f(\mathbf{x_i};\theta)$ \tcp*{Sec~\ref{sec:enc}}
    Embedding-Level augmentation and label sequence decoding $\mathbf{\hat{y}_i}=g(\mathbf{z_i};\phi)$ \tcp*{Sec~\ref{sec:dec},\ref{sec:pretrain-aug}}
    Optimize $\theta$ and $\phi$ through Equation~\ref{eq:ce};
    }

    \For{$(\mathbf{x_i, y_i}) \in \mathcal{D}_{\mathrm{te}}$}{
    Encoder feature extraction $\mathbf{z_i}=f(\mathbf{x_i};\theta)$;
    
    Embedding-Level augmentation and label sequence constrained decoding $\mathbf{\hat{y}_{i}}=\mathrm{argmax}_{\mathbf{y_i} \in \mathcal{Y}} \ P_{\phi}(\mathbf{y_i}|\mathbf{z_i})$;
    }

    \Return predicted label sequences $\{\mathbf{\hat{y}_{i}}\}_{i=0}^{N_{\mathrm{te}}}$.

  \caption{\our Framework}
  \label{alg}
\end{algorithm}

\section{Methodology}\label{sec:method}

We design a label structure decoding architecture for HAR, called \our, that exploits label structures and promotes knowledge sharing across activities. 
\our consists of two modules: Time-Series Encoder and Label Structure-Constrained Decoder. We pass multivariate sensory readings as input to the encoder and use the extracted feature vector to initialize the hidden states of the decoder. The decoder generates an activity name sequence (e.g., ``climb stairs'') as the prediction label. By binding sensory features with label structures, the structures in label names help the model better learn the similarity in the sensory data. We further propose three augmentation methods, including one basic token-level augmentation (randomly selecting from ``climb'', ``stairs'', ``climb stairs'') and two pre-trained model-enhanced augmentations at embedding (using pre-trained embeddings to initialize ``climb'' and ``stairs'' word embeddings) and label sequence levels (rephrasing ``climb stairs'' as ``leg up'' which share more tokens with other label names), to better capture shared structures across different activities. We summarize the pipeline of \our in Figure~\ref{fig:model} and Algorithm~\ref{alg}.

\subsection{Time-Series Encoder}\label{sec:enc}
We use $f_{\theta} : \mathcal{X} \rightarrow \mathcal{Z} \subset \mathbf{R}^{d}$ parameterized by $\theta$ to denote the time-series encoder. This part appears in both conventional HAR and \our. The encoder maps data from the input space $\mathcal{X}$ to the $d$-dimensional hidden space $\mathcal{Z}$. For conventional HAR, the final predictions are obtained from the hidden representations after a fully connected layer $\mathrm{fc}$. Denote $\hat{c}_i = \mathrm{fc}(f(\mathbf{x_i};\theta)) \in \mathbf{R}^{C}$ as the distribution of the predicted label. Optimization is based on the cross-entropy loss between prediction $\hat{c}_i$ and ground truth $c_i$:
\begin{equation}
    \mathop{\arg\min} \mathbb{E}_{(\mathbf{x_i}, c_i)\sim \mathcal{D'}} \mathrm{CE}(c_i, \hat{c}_i).
\end{equation}

In \our, the encoded representations $\mathbf{z_i}=f(\mathbf{x_i};\theta)$ are used to initialize hidden states of the decoder, instead of being directly used for classification. This transfers learned representations from the encoder to inform the structured decoding process. To instantiate the time-series encoder, we keep both efficacy and efficiency in mind, given that HAR models usually run on edge devices with limited compute. Therefore, we use one-dimensional Convolutional Neural Networks (CNN), as they are relatively lightweight with superior capability in extracting time-series features~\cite{wang2017time,cui2016multi,radu2018multimodal,zhang2022esc}. 

\subsection{Label Structure-Constrained Decoder}\label{sec:dec}

We use $g_{\phi}: \mathcal{Z} \rightarrow \mathcal{Y}$ parameterized by $\phi$ to denote the label structure-constrained decoder in \our. The decoder generates word sequences in the label space $\mathcal{Y}$ given the encoded representations as initialization of the decoder hidden states. Following our notation in Section~\ref{sec:setting} (Problem Setting), we further require that each label name sequence starts from a start token \textlangle$s$\textrangle \ and ends at an ending token \textlangle$e$\textrangle. Specifically, $\mathbf{y_i}=[y_{i0},y_{i1},y_{i2},\cdots,y_{ik_i},y_{ik_i+1}]$, where $y_{i0} = \ $\textlangle$s$\textrangle, $y_{ik+1} = \ $\textlangle$e$\textrangle. Decoding the token \textlangle$e$\textrangle \ means that we reach the end of the label sequence. At each decoding step, we estimate the conditional probability $P_{\phi}$ of decoding label $\mathbf{y_i}$ from $\mathbf{x_i}$, given the encoded representations $\mathbf{z_i}$ from the encoder as:
\begin{equation}
\resizebox{.91\linewidth}{!}{$
    P_{\phi}(y_{i1},y_{i2},\cdots,y_{ik_i+1}|\mathbf{z_i}) = \prod_{t=1}^{k_i+1} P_{\phi}(y_{it}|\mathbf{z_i},y_{i0},y_{i1},\cdots,y_{it-1}).
$}
\end{equation}

\noindent \textbf{Training.} During the training of \our, we adopt the teacher forcing strategy~\cite{williams1989learning} where the ground truth label token $y_{it}$ at each decoding step $t$ is used as input to be conditioned on for predictions at decoding step $t+1$. Teacher forcing improves convergence speed and stability during training. We optimize \our based on cross-entropy loss between the predicted label sequence $\mathbf{\hat{y}_i}$ and the ground truth label  sequence $\mathbf{y_i}$: 
\begin{equation}\label{eq}
\mathbf{\hat{y}_i} = g(f(\mathbf{x_i};\theta);\phi),
\end{equation}
\begin{equation}
    \mathop{\arg\min} \mathbb{E}_{(\mathbf{x_i}, \mathbf{y_i})\sim \mathcal{D}}\frac{1}{k_i}\sum_{j=1}^{k_i} \mathrm{CE}(y_{ij},\hat{y}_{ij}),
    \label{eq:ce}
\end{equation}
where $\hat{y}_{ij}\in \mathbf{R}^{M}$ indicates distribution of $j$th predicted token of $\mathbf{\hat{y}_{i}}$.

\noindent \textbf{Inference with Constrained Decoding.} During inference decoding, predicted label token $\hat{y}_{it}$ from the current decoding step $t$ is used as input to be conditioned on for predicting tokens at step $t+1$. 
In typical natural language processing tasks, e.g., machine translation, it is common to decode the sequence using beam search during inference. However, 
beam search would not work properly as it only tracks a pre-defined number of best partial solutions as candidates in decoding, and the final predictions may not belong to our label space. To guarantee that all generated labels are valid, we adopt a \emph{constrained decoding} method. We start from the start token and iterate over all valid label sequences in the label set. We then calculate the probability of decoding each valid label sequence and choose the one with the highest probability as the final predicted label.
The decoding is constrained as we only keep track of the valid partial sequences during decoding. 
In HAR datasets, the size of the label set is relatively small, and constrained decoding consumes only a small constant of memory (the size of the label set). At step $t$, we calculate the probability for all the valid partial sequences of length $t$ and pass them into the decoder for generating tokens at step $t+1$. The final inference prediction is the sequence that maximizes the overall sequence probability:
\begin{equation}\label{eq:inference}
    \mathbf{\hat{y_i}} = \mathop{\arg\max}_{\mathbf{y_i} \in \mathcal{Y}} P_{\phi}(\mathbf{y_i}|\mathbf{z_i}).
\end{equation}
We use Long Short-Term Memory (LSTM) as an example for our label structure-constrained decoder, given its effectiveness in modeling sequential dependencies~\cite{ordonez2016deep}. We transform the CNN-extracted features $\mathbf{z_i}$ through two separate linear layers to initialize the hidden state and cell state of LSTM. 

\subsection{Basic Token-Level Label Augmentation}\label{sec:token-aug}
\begin{figure}[t]
\centering
\includegraphics[width=0.9\linewidth]{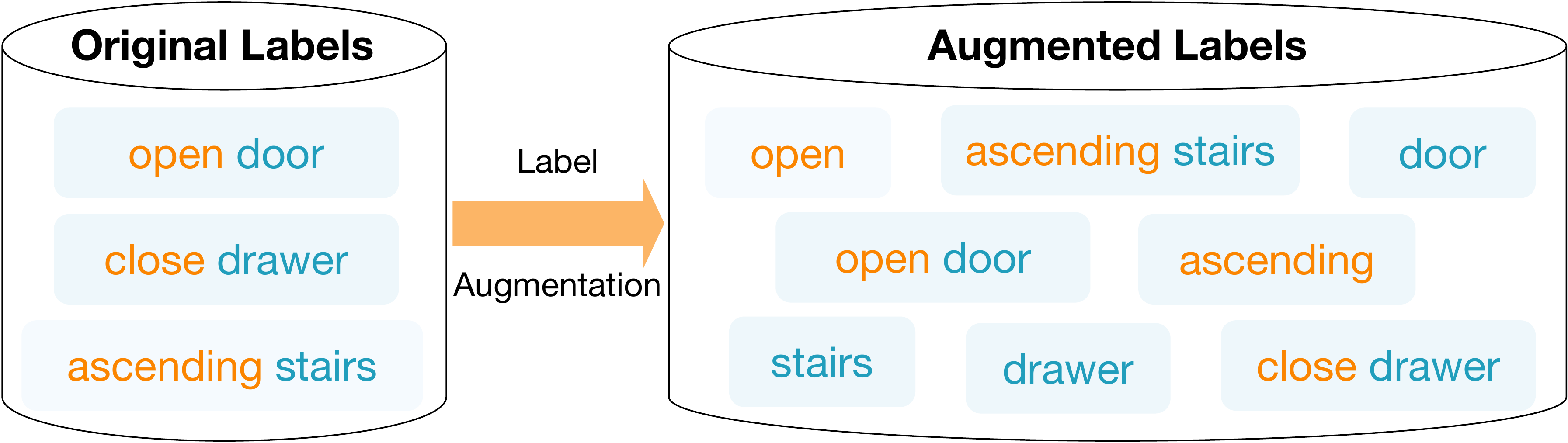}
\vspace{-2mm}
\caption{Illustration of basic token-level augmentation. We augment the original label name sequence by randomly choosing its meaningful tokens 
or the sequence itself 
.}
\label{fig:aug}
\vspace{-3mm}
\end{figure}

To better learn the semantics of each token in the label sequence, we apply a token-level label augmentation strategy as illustrated in Figure~\ref{fig:aug}. During training, with pre-defined probability, we randomly choose meaningful single words from the original label sequence as the new labels. For example, an original label sequence ``ascending stairs'' contains single words ``ascending'' and ``stairs'', so we randomly select from ``ascending'', ``stairs'', and ``ascending stairs'' as the new labels during training. Following notation in Section~\ref{sec:setting} (Problem Setting), the original label $\mathbf{y_i}=[y_{i1}, y_{i2}, \cdots, y_{ik_i}]$ is augmented as a set of new labels $\{\mathbf{y_i}, y_{i1}, y_{i2}, \cdots, y_{ik_i}\}$ containing the label sequence $\mathbf{y_i}$ and its meaningful tokens. For each iteration, with a pre-defined probability we randomly select the new label $\mathbf{y'_i}$ from the new label set as the actual label. Optimization with token-level label augmentation can be formulated as:
\begin{equation}
    \mathop{\arg\min} \mathbb{E}_{(\mathbf{x_i}, \mathbf{y_i})\sim D}  \mathbb{E}_{\mathbf{y'_i}\sim \{\mathbf{y_i}, y_{i1}, y_{i2}, \cdots, y_{ik_i}\}}\frac{1}{k'_i}\sum_{j=1}^{k'_i} \mathrm{CE}(y'_{ij},\hat{y}_{ij}),
\end{equation}
where $k'_i$ is the length of the new label $\mathbf{y'_i}$, $y'_{ij}$ is the $j$th token of $\mathbf{y'_i}$, and $\hat{y}_{ij}$ is the distribution of the predicted $j$th token. Since the goal of label augmentation is to help the model better capture the semantics of different activities, we only choose meaningful single tokens in the original label sequences (e.g., actions and objects) as new labels. Other single tokens like stop words or numbers (e.g., ``1'' in ``open door 1'') will not count as new labels. Note that the token-level augmentation is only applied during training. During evaluation, the ground truth label stays the same as the original label. Because we adopt a constrained decoding method during inference, it is guaranteed that all the generated label sequences are valid sequences in the original label sets. 

\begin{table*}[t]
\centering
\small
    \caption{Dataset statistics and an example subset of shared label names.}
    \vspace{-3mm}
	\setlength{\tabcolsep}{1.1mm}{
    \begin{tabular}{r|cccccc}
    \hline
        Dataset& Train & Test & Window Size & Channel & Class Num & An Example Subset of Shared Label Names \\
    \hline
    \hline
        Opportunity~\cite{roggen2010collecting} & 2891 & 235 & 150 & 45 & 17 & open door, open drawer, close drawer, open fridge, open dishwasher \\
        
        PAMAP2~\cite{reiss2012introducing} & 14438 & 2380  & 512 & 27 & 12 & ascending stairs, descending stairs, walking, nordic walking \\

        UCI-HAR~\cite{anguita2013public} & 7352 & 2947 & 128 & 9 & 6 & walk, walk upstairs, walk downstairs \\

        USCHAD~\cite{zhang2012usc} &17576 & 9769&100 & 6& 12& run forward, walk forward, elevator up, elevator down, jump up\\

        WISDM~\cite{weiss2019smartphone} &12406 & 3045&200 &6 &18 & eating soup, eating pasta, kicking soccer ball, playing tennis ball \\

        Harth~\cite{logacjov2021harth} &14166 & 3588& 300 &6 &12 & sitting, standing, cycling sitting, cycling standing, cycling sitting inactive \\
 
    \hline
    \end{tabular}}
    \label{tab:dataset}
\end{table*}

\begin{table*}[t]
\centering
\small
\caption{Accuracy and Macro-F1 for \our and baselines. We \textbf{bold} the best score and \underline{underline} the second best. }
\vspace{-3mm}
\scalebox{0.87}{
\begin{tabular}{l|c|ccccccccc}
\toprule
Datasets         & Metrics  & DeepConvLSTM~\cite{ordonez2016deep}      & XGBoost~\cite{chen2016xgboost}  & MA-CNN~\cite{radu2018multimodal} & HHAR-net~\cite{fazli2021hhar} & TST~\cite{zerveas2021transformer} & TARNet~\cite{chowdhury2022tarnet}  & Rocket~\cite{dempster2020rocket} & THAT~\cite{li2021two} & \our \\ \hline \hline 
\multirow{2}{*}{Opp} & Accuracy & 0.746$\pm$0.049 & 0.688$\pm$0.017 & 0.549$\pm$0.029 & 0.753$\pm$0.027      & 0.784$\pm$0.018 & 0.789$\pm$0.024 &  \underline{0.811$\pm$0.008} & 0.803$\pm$0.012 & \textbf{0.849$\pm$0.015} \\
                           & Macro-F1 & 0.634$\pm$0.036 & 0.547$\pm$0.011 &  0.416$\pm$0.036 & 0.620$\pm$0.021 & 0.668$\pm$0.023 & 0.669$\pm$0.034 & 0.670$\pm$0.016 & \underline{0.691$\pm$0.015} &   \textbf{0.766$\pm$0.013}   \\ \hline
\multirow{2}{*}{PAMAP2}      & Accuracy & 0.891$\pm$0.012 & 0.939$\pm$0.003 &  0.926$\pm$0.011      &    0.885$\pm$0.031 & 0.922$\pm$0.037 & 0.931$\pm$0.011 &   0.928$\pm$0.008 & \underline{0.943$\pm$0.005} &  \textbf{0.960$\pm$0.002}                 \\
                             & Macro-F1 & 0.884$\pm$0.018 & 0.939$\pm$0.007 &  0.925$\pm$0.012      &   0.893$\pm$0.031  & 0.925$\pm$0.039 & 0.935$\pm$0.010 & 0.934$\pm$0.008  &   \underline{0.949$\pm$0.005}   &    \textbf{0.965$\pm$0.002}                  \\ \hline
\multirow{2}{*}{UCI-HAR}      & Accuracy & 0.900$\pm$0.016 & 0.907$\pm$0.003 &  0.921$\pm$0.025      & 0.926$\pm$0.005 & 0.926$\pm$0.005 & 0.904$\pm$0.011 &   \underline{0.939$\pm$0.002} & 0.906$\pm$0.007 &  \textbf{0.960$\pm$0.002}                \\
                             & Macro-F1 & 0.899$\pm$0.016 & 0.906$\pm$0.003 &  0.921$\pm$0.024      &   0.926$\pm$0.005    & 0.925$\pm$0.006   & 0.904$\pm$0.011 &    \underline{0.942$\pm$0.002}   & 0.909$\pm$0.006    & \textbf{0.959$\pm$0.002}                  \\ \hline
\multirow{2}{*}{USCHAD}      & Accuracy & 0.574$\pm$0.016 & 0.571$\pm$0.007 &  0.543$\pm$0.044      &   0.524$\pm$0.011  & 0.641$\pm$0.028 & 0.564$\pm$0.037  &   0.580$\pm$0.005   & \underline{0.643$\pm$0.015}    &  \textbf{0.674$\pm$0.041}                 \\
                             & Macro-F1 & 0.557$\pm$0.015 & 0.573$\pm$0.006 &  0.520$\pm$0.047      &   0.523$\pm$0.009     &  0.594$\pm$0.023 & 0.533$\pm$0.021  &  0.601$\pm$0.007 & \underline{0.619$\pm$0.012}  &  \textbf{0.627$\pm$0.027}             \\ \hline
\multirow{2}{*}{WISDM}       & Accuracy & 0.689$\pm$0.014 & 0.668$\pm$0.005 & 0.634$\pm$0.059 & 0.566$\pm$0.016 & 0.715$\pm$0.003 & 0.733$\pm$0.011 & 0.643$\pm$0.007 & \underline{0.774$\pm$0.005} & \textbf{0.794$\pm$0.003}                  \\
                             & Macro-F1 & 0.685$\pm$0.013 & 0.662$\pm$0.006 &  0.631$\pm$0.060      &   0.538$\pm$0.012     & 0.710$\pm$0.004    & 0.737$\pm$0.010 &    \underline{0.767$\pm$0.004 }       & 0.634$\pm$0.005    &  \textbf{0.790$\pm$0.004}                \\ \hline
\multirow{2}{*}{Harth}       & Accuracy & 0.979$\pm$0.006 & 0.977$\pm$0.001 &  0.973$\pm$0.016      &   \underline{0.981$\pm$0.001}     & 0.974$\pm$0.005   & 0.962$\pm$0.009   & 0.897$\pm$0.003  & 0.960$\pm$0.016     &    \textbf{0.983$\pm$0.007}              \\
                             & Macro-F1 & \underline{0.578$\pm$0.032} & 0.522$\pm$0.003 &  0.538$\pm$0.025      &   0.515$\pm$0.049     &  0.501$\pm$0.031    & 0.481$\pm$0.031 & 0.472$\pm$0.019      & 0.485$\pm$0.025    &    \textbf{0.593$\pm$0.020}   \\     
\bottomrule
\end{tabular}
}
\label{tab:result}
\end{table*}

\subsection{Enhanced Embedding-Level and Sequence-Level Augmentations}\label{sec:pretrain-aug}

Apart from the basic token-level augmentation, we also develop two enhanced augmentation techniques to better capture label structures from embedding and sequence levels by leveraging the power of pre-trained models. 

\begin{figure}[t]
\centering
\begin{subfigure}[c]{0.235\textwidth}

  \centering
  \includegraphics[width=\textwidth]{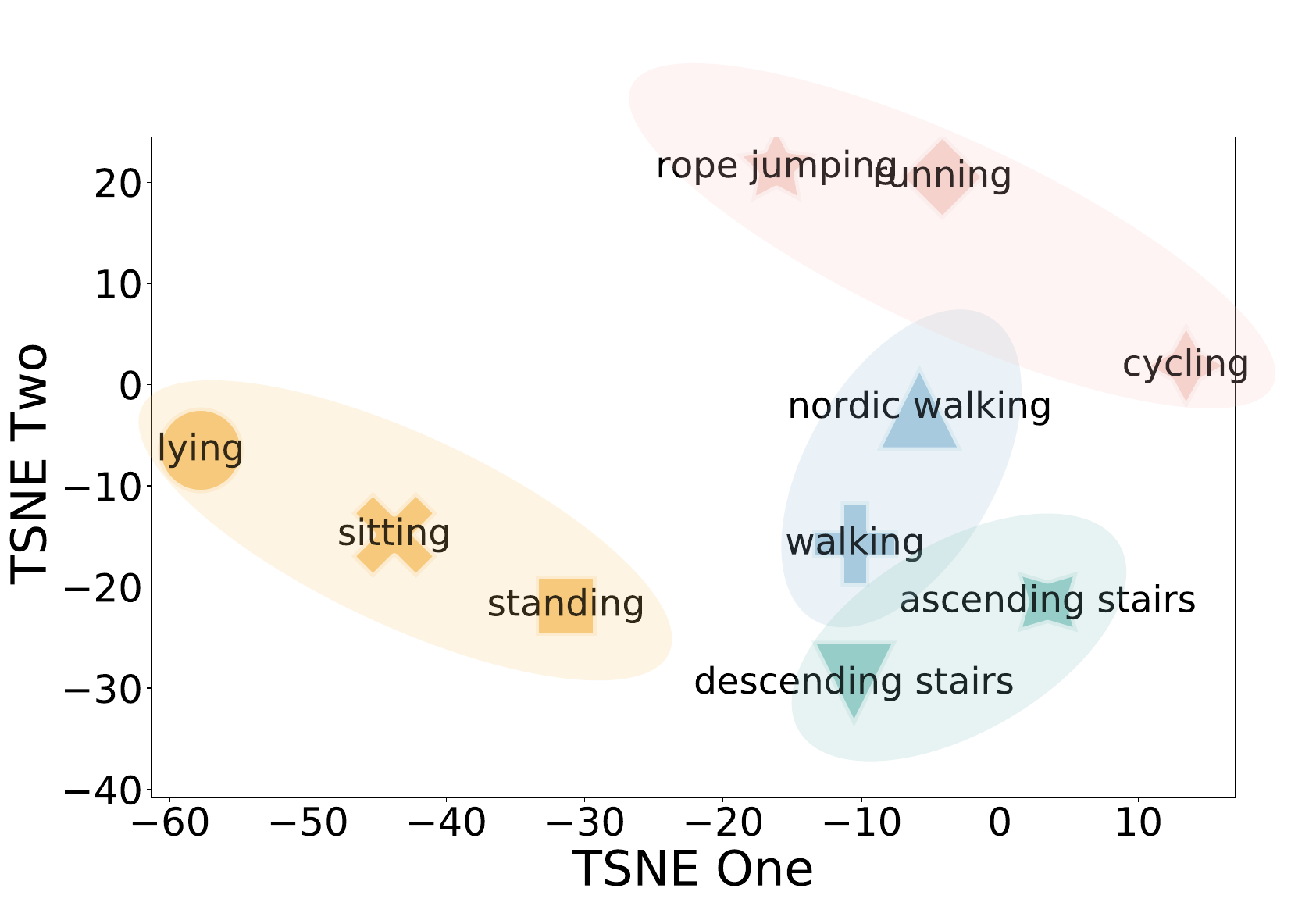}
  \vspace{-5mm}
  \caption{Data}
\end{subfigure}
\begin{subfigure}[c]{0.235\textwidth}
  \centering
  \includegraphics[width=\textwidth]{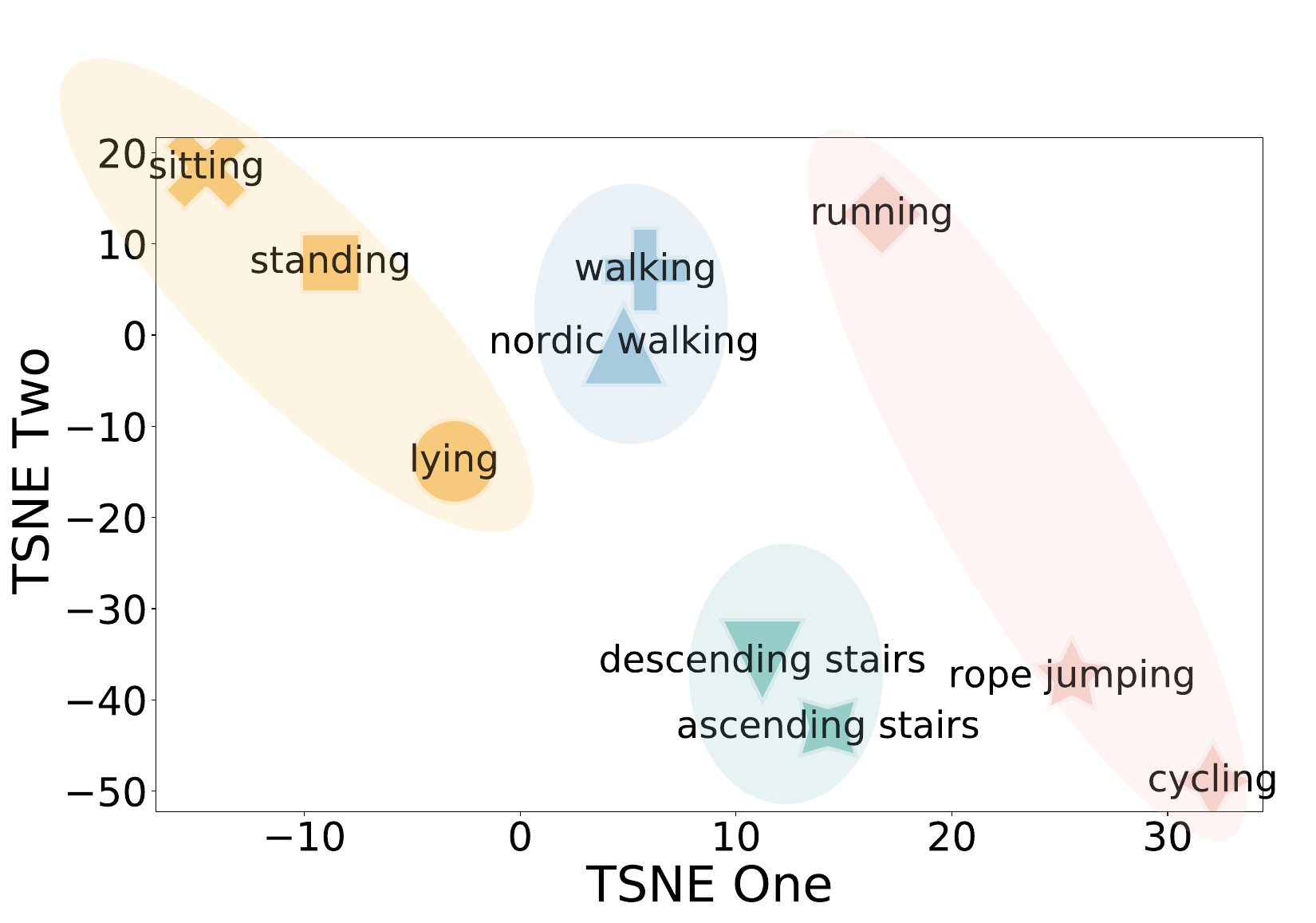}
  \vspace{-5mm}
  \caption{ImageBind Embeddings}
\end{subfigure}
\vspace{-3mm}
\caption{T-SNE visualizations show analogous clusters between input data and ImageBind word embeddings. Activities in the same color represent clusters of similar activities.}
\label{fig:imagebind}
\vspace{-5mm}
\end{figure}

\noindent \textbf{Embedding-Level Augmentation.} Our label structure decoding architecture can capture label structures explicitly presented as shared label names. Yet, apart from these explicit shared label names, there may also exist semantic structures that implicitly span across different activities. For example, ``walk'' and ``run'' are similar activities involving the movement of legs, but they don't directly share label names. We have observed that such semantic structures can be captured by word embeddings from pre-trained models. We thus propose to use word embeddings from pre-trained models to initialize our decoder's word embedding layer, replacing the original random initialization. Specifically, we utilize word embeddings from ImageBind, a multimodal pre-trained model that learns a joint embedding space across six modalities.
As shown in Figure~\ref{fig:imagebind}, we apply t-SNE visualization to both the ImageBind word embeddings and the input sensor readings from some example activities in PAMAP dataset~\cite{reiss2012introducing}. For activity names comprising multiple tokens, we calculate the average embedding of the aggregated tokens. T-SNE visualizations show similar clusters between ImageBind word embeddings and original data embeddings. As a result, incorporating pre-trained word embeddings helps \our better capture semantic structures. 

\noindent \textbf{Sequence-Level Augmentation.} Most HAR datasets have sufficient overlapping structures in label names. However, there also exist datasets that do not have or rarely have shared tokens in their original label names. For these datasets, we can use large-scale language models to automatically generate label names with shared tokens. Specifically, we employ GPT-4 with the following prompt: 

\textit{Describe the following activities one by one with information of 1. body part used, 2. action or adverb, 3. object (if involved). Please maximize the number of shared tokens across different activities and make the description as short as possible.}

As human activities naturally have shared actions and objects, the
prompt helps find common tokens across activities. With the aid of
pre-trained language model, such a process is performed with minimal human expert effort. Based on the structured information provided by the pre-trained model, we can summarize the label names with shared tokens. We apply sequence-level augmentation mostly for datasets without original shared tokens. If the target HAR dataset already has sufficient overlapping tokens, we will directly use the original label names provided by human experts.

\section{Evaluation}

\subsection{Datasets, Baselines, and Metrics}

We use six HAR benchmark datasets for evaluation, summarized in Table~\ref{tab:dataset} with examples of shared label names. We split data and choose window size following previous works~\cite{jeyakumar2019sensehar,chowdhury2022tarnet}. The training and testing split is based on different participating subjects.

\noindent \textbf{Opportunity}\footnote{\url{https://archive.ics.uci.edu/ml/datasets/opportunity+activity+recognition}}~\cite{roggen2010collecting} collects readings from 4 users with 6 runs per user. Sensors include body-worn, object, and ambient sensors. The full dataset includes annotations on multiple levels, and we use mid-level gesture annotations which contain shared label structures.  

\noindent \textbf{PAMAP2}\footnote{\url{http://archive.ics.uci.edu/ml/datasets/pamap2+physical+activity+monitoring}}~\cite{reiss2012introducing} comprises readings collected from 9 subjects wearing 3 IMUs sampled at 100 Hz and a heart rate monitor sampled at 9Hz. Three IMUs are positioned over the wrist on the dominant arm, on the chest, and on the dominant side's ankle, respectively. 

\noindent \textbf{UCI-HAR}\footnote{\url{http://archive.ics.uci.edu/ml/datasets/Human+Activity+Recognition+Using+Smartphones}}~\cite{anguita2013public} is collected from a group of 30 volunteers. A Samsung Galaxy S II smartphone was attached on their waist. Feature vectors were further extracted from each sliding window of the collected data in the time and frequency domain. 

\noindent \textbf{USCHAD}\footnote{\url{https://sipi.usc.edu/had/}}~\cite{zhang2012usc} involves 14 subjects performing 12 low-level activities. They use MotionNode (6-DOF IMU designed for human motion sensing applications) to collect the datasets. 

\noindent \textbf{WISDM}\footnote{\url{https://archive.ics.uci.edu/ml/datasets/WISDM+Smartphone+and+Smartwatch+Activity+and+Biometrics+Dataset+}}~\cite{weiss2019smartphone} is collected from accelerometer and gyroscope sensors in smartphone and smartwatch at a rate of 20Hz. 51 subjects perform 18 activities for 3 minutes respectively. 

\noindent \textbf{Harth}\footnote{\url{https://github.com/ntnu-ai-lab/harth-ml-experiments}}~\cite{logacjov2021harth} involves 22 subjects using two three-axial accelerometers attached to the thigh and lower back, and a chest-mounted camera (for data annotation) to collect data of 12 activities.

\begin{table*}[t]
\centering
\small
\caption{Accuracy and Macro-F1 for different model variants. We \textbf{bold} the best score and \underline{underline} the second best.}
\vspace{-3mm}
\scalebox{0.95}{
\begin{tabular}{l|c|cccccc|c|c}
\toprule 
\multirow{2}{*}{Datasets}                    & \multirow{2}{*}{Metrics} & \van & VanillaHAR+ & \multirow{2}{*}{multi label} &\multirow{2}{*}{no aug}&\multirow{2}{*}{no token aug} &\multirow{2}{*}{no embed aug} &  \multirow{2}{*}{best baseline}      & \multirow{2}{*}{\our}           \\ 
 & & no label modeling & ImageBind & & & & \\ \hline \hline
\multirow{2}{*}{Opp} & Accuracy & 0.745$\pm$0.015 & 0.759$\pm$0.010 & 0.755$\pm$0.030  &0.819$\pm$0.005& {0.823$\pm$0.022} 
&\underline{0.847$\pm$0.015} & 0.811$\pm$0.008 & \textbf{0.849$\pm$0.015}               \\
                             & Macro-F1 & 0.618$\pm$0.023 & 0.632$\pm$0.009 & 0.624$\pm$0.034 &0.732$\pm$0.014 &{0.737$\pm$0.019} & \underline{0.741$\pm$0.029} & 0.691$\pm$0.015 &\textbf{0.766$\pm$0.013}                  \\ \hline
\multirow{2}{*}{PAMAP2}      & Accuracy &  0.921$\pm$0.033 & 0.933$\pm$0.011 & 0.926$\pm$0.011 & 0.951$\pm$0.006& {0.952$\pm$0.005} & \underline{0.956$\pm$0.008} & 0.943$\pm$0.005 &\textbf{0.960$\pm$0.002}            \\
                             & Macro-F1 & 0.924$\pm$0.024 & 0.938$\pm$0.014 & 0.928$\pm$0.010  & 0.960$\pm$0.006&{0.958$\pm$0.005} & \underline{0.964$\pm$0.009} & 0.949$\pm$0.005 &  \textbf{0.965$\pm$0.002}                  \\ \hline
\multirow{2}{*}{UCI-HAR}      & Accuracy & 0.921$\pm$0.005 & 0.928$\pm$0.005 & 0.926$\pm$0.005 & 0.954$\pm$0.006&{0.957$\pm$0.004} & \underline{0.958$\pm$0.004} & 0.939$\pm$0.002&\textbf{0.960$\pm$0.002}                \\
                             & Macro-F1 &  0.921$\pm$0.005 & 0.928$\pm$0.005 & 0.926$\pm$0.005 & 0.953$\pm$0.006&{0.957$\pm$0.004}& \underline{0.958$\pm$0.004} & 0.942$\pm$0.002&\textbf{0.959$\pm$0.002}                  \\ \hline
\multirow{2}{*}{USCHAD}      & Accuracy & 0.543$\pm$0.028 & 0.566$\pm$0.013 & 0.550$\pm$0.023 & 0.622$\pm$0.050& 0.635$\pm$0.034& \underline{0.663$\pm$0.012} &{0.643$\pm$0.015}&\textbf{0.674$\pm$0.041}                 \\
                             & Macro-F1 & 0.538$\pm$0.024 & 0.558$\pm$0.012 & 0.546$\pm$0.016 & 0.600$\pm$0.029& 0.615$\pm$0.023 & \underline{0.623$\pm$0.011} &{0.619$\pm$0.012}&\textbf{0.627$\pm$0.027}             \\ \hline
\multirow{2}{*}{WISDM}       & Accuracy  &  0.644$\pm$0.006 & 0.655$\pm$0.004 & 0.649$\pm$0.010 &   {0.788$\pm$0.006} & 0.786$\pm$0.008&\underline{0.793$\pm$0.008} & 0.774$\pm$0.005 &  \textbf{0.794$\pm$0.003}            \\
                             & Macro-F1 &   0.639$\pm$0.006 & 0.648$\pm$0.003  & 0.645$\pm$0.012 & 0.783$\pm$0.008&  {0.784$\pm$0.008} & \underline{0.786$\pm$0.009} &0.767$\pm$0.004 &     \textbf{0.790$\pm$0.004}         \\ \hline
\multirow{2}{*}{Harth}       & Accuracy &  0.977$\pm$0.005 & 0.980$\pm$0.002 & 0.979$\pm$0.007 & 0.981$\pm$0.006 & {0.975$\pm$0.008} & \underline{0.982$\pm$0.003} & {0.981$\pm$0.001} & \textbf{0.983$\pm$0.007}   \\
                             & Macro-F1  &  0.481$\pm$0.004 & 0.487$\pm$0.014 & 0.482$\pm$0.005  &0.566$\pm$0.034&\underline{0.591$\pm$0.043} & 0.583$\pm$0.023 & {0.578$\pm$0.032} & \textbf{0.593$\pm$0.020}   \\
\bottomrule
\end{tabular}
}
\label{tab:abl}
\end{table*}

\begin{table*}[t]
\centering
\small
\caption{Accuracy and Macro-F1 on Mhealth dataset. We \textbf{bold} the best score and \underline{underline} the second best.}
\vspace{-3mm}
\scalebox{0.86}{
\begin{tabular}{l|c|ccccccccc}
\toprule
Datasets         & Metrics  & DeepConvLSTM~\cite{ordonez2016deep}      & XGBoost~\cite{chen2016xgboost}  & MA-CNN~\cite{radu2018multimodal} & HHAR-net~\cite{fazli2021hhar} & TST~\cite{zerveas2021transformer} & TARNet~\cite{chowdhury2022tarnet}  & Rocket~\cite{dempster2020rocket} & THAT~\cite{li2021two} & \our \\ \hline \hline 
\multirow{2}{*}{Mhealth} & Accuracy &  0.868$\pm$0.023 & 0.809$\pm$0.011 & 0.839$\pm$0.010  & 0.854$\pm$0.020 & 0.863$\pm$0.007 & 0.895$\pm$0.042 & 0.902$\pm$0.006  & \underline{0.907$\pm$0.019}  &  \textbf{0.975$\pm$0.014}      \\
                             & Macro-F1 & 0.871$\pm$0.023 & 0.775$\pm$0.023 & 0.834$\pm$0.009  & 0.811$\pm$0.022 & 0.863$\pm$0.007  & 0.892$\pm$0.039 &  0.908$\pm$0.007 & \underline{0.910$\pm$0.017} &  \textbf{0.974$\pm$0.013}          \\     
\bottomrule
\end{tabular}
}
\label{tab:mhealth}
\end{table*}

\begin{table}[t]
\centering
\small
\caption{Different model variants on Mhealth dataset. We \textbf{bold} the best score and \underline{underline} the second best.}
\vspace{-3mm}
\scalebox{0.83}{
\begin{tabular}{l|c|cccc}
\toprule
Datasets         & Metrics  & no token aug & no embed aug & no seq aug & \our \\ \hline \hline 
\multirow{2}{*}{Mhealth} & Accuracy & \underline{0.968$\pm$0.027} &     0.949$\pm$0.019 & 0.908$\pm$0.008 & \textbf{0.975$\pm$0.014} \\
    & Macro-F1 & \underline{0.969$\pm$0.027} &  0.949$\pm$0.021   & 0.905$\pm$0.012 & \textbf{0.974$\pm$0.013}\\     
\bottomrule
\end{tabular}
}
\vspace{-3mm}
\label{tab:mhealth-abl}
\end{table}

We compare \our with a list of human activity recognition (DeepConvLSTM~\cite{ordonez2016deep}, MA-CNN~\cite{radu2018multimodal}, HHAR-net~\cite{fazli2021hhar}, THAT~\cite{li2021two}) and time-series classification baselines (XGBoost~\cite{chen2016xgboost}, Rocket~\cite{dempster2020rocket}, TST~\cite{zerveas2021transformer}, TARNet~\cite{chowdhury2022tarnet}), including both statistical approaches and state-of-the-art deep learning-based models. 

We evaluate the performance of \our and baselines using accuracy and macro-F1. Macro-F1 is defined as macro-F1$ = \frac{1}{C} \sum_{i=1}^C 2 \times \frac{\mathrm{Prec_i} \times \mathrm{Rec_i}}{\mathrm{Prec_i + Rec_i}}$, where $\mathrm{Prec_i,Rec_i}$ represent the precision and recall for each category $i$, and $C$ is the total number of categories.

\vspace{-3mm}
\subsection{Experimental Setup}
We use a two-layer convolutional neural network as the encoder for extracting features. The kernel sizes for both layers are set to $3$ and each layer is followed by batch normalization. We adopt LSTM with a hidden dimension of $128$ as the decoder, based on a grid search of $\{64,128,256\}$. 
We use Adam optimizer with learning rate $1e^{-4}$ based on a grid search of $\{1e^{-5},1e^{-4},1e^{-3},1e^{-2}\}$ and batch size $16$. For all datasets, we further randomly split the training set into 80\% for training and $20\%$ for validation. We conduct the experiments in \textsc{Pytorch} with NVIDIA RTX A6000 (with 48GB memory), AMD EPYC 7452 32-Core Processor, and Ubuntu 18.04.5 LTS. We tune the hyper-parameters of both \our and baselines on the validation set and then combine training and validation set to re-train the models after hyper-parameter tuning. 

\begin{table}[t]
\centering
\small
\caption{Original/generated label names for Mhealth data.}
\vspace{-3mm}
\begin{tabular}{l|l}
\toprule
Original Label Names                     & Generated Label Names \\ \hline \hline 
standing still & leg still         \\  
sitting and relaxing & buttocks still \\
lying down & back down \\
walking & leg walk \\
climbing stairs & leg up \\
waist bends forward & back forward \\
frontals elevation of arms & arm up \\
knees bending (crouching) & leg forward \\
cycling & leg cycle \\
jogging & leg jog \\
running & leg jog fast \\
jump front and back & leg jump \\
\bottomrule
\end{tabular}
\vspace{-3mm}
\label{tab:mhealth-label-names}
\end{table}

\subsection{Results}
We repeat $5$ runs and report the average accuracy, macro-F1 score, and standard deviations of \our and baselines in Table~\ref{tab:result}. We see that \our consistently outperforms both statistical and deep learning-based human activity recognition and time-series classification approaches, in terms of both accuracy and macro-F1 score. \our reduces the error rate (i.e., 1 - accuracy) on six datasets by approximately 20\%, 30\%, 34\%, 9\%, 9\%, 11\% compared with each dataset's best-performing baseline. 
Compared with the hierarchical baseline HHAR-net which models activities in a simple 2-layer hierarchical model, \our can model much more complex dependencies not necessarily in a hierarchical structure (e.g., ``open door'', ``open drawer'', ``close drawer'' with pairwise overlap, forming a graph rather than tree structure), without the cost of manual labeling from experts. 
TST and TARNet leverage unsupervised representation learning to boost classification performance. However, they do not explicitly take account of label structures to model relations across different activities. 
Other top-performing HAR or time-series classification methods, such as Rocket and THAT, propose better feature extractors to improve recognition performance, but they also neglect the label name structures. \our is capable of leveraging the inherent shared structures in label names, leading to the highest accuracy and macro-F1 score. 

To assess the statistical significance of the performance differences between \our and the baselines, we applied 
%the Friedman test and 
the Wilcoxon-signed rank test with Holm's $\alpha$ (5\%) following the procedures described in ShapeNet~\cite{li2021shapenet,holm1979simple}.  
The Wilcoxon-signed rank test indicates that the improvement of \our compared with all the baselines is statistically significant with $p$ far below 0.05 (e.g., $p=5e^{-4}$ for the best-performing baseline THAT).

\begin{figure*}[t]
      \centering
      \begin{subfigure}[b]{0.49\textwidth}
          \centering
          \includegraphics[width=\textwidth]{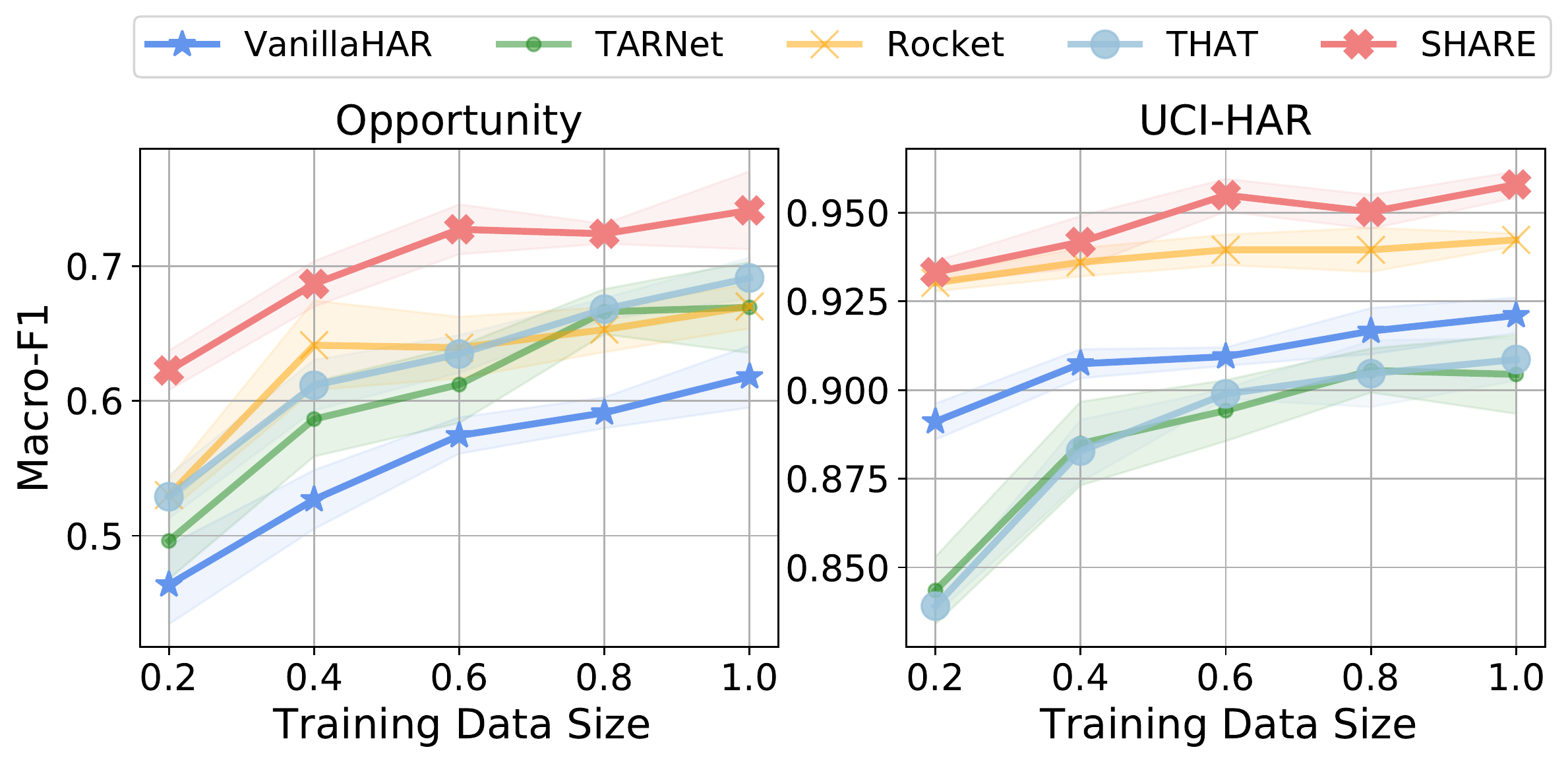}
          \caption{Reduced Training Sample}
          \label{fig:ds}
      \end{subfigure}
      \begin{subfigure}[b]{0.49\textwidth}
          \centering
          \includegraphics[width=\textwidth]{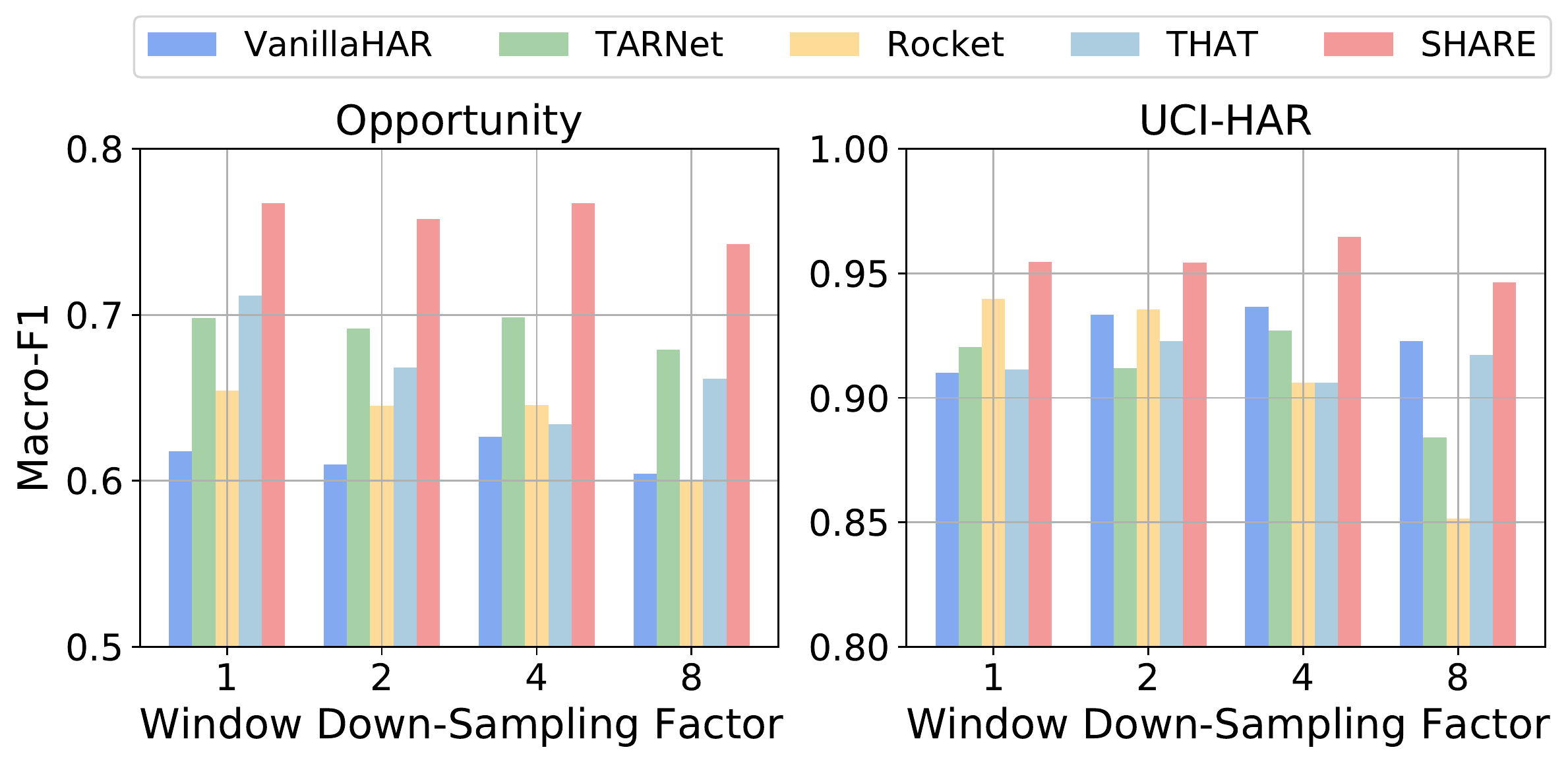}
          \caption{Reduced Window Size}
          \label{fig:sam}
      \end{subfigure}
         \label{fig:robustness}
     \vspace{-3mm}
      \caption{Macro-F1 of \our, \van and best-performing baselines with reduced training samples and window size.}
      \vspace{-3mm}
\end{figure*}

\begin{figure}[t]
      \centering
      \begin{subfigure}[b]{0.23\textwidth}
          \centering
          \includegraphics[width=\textwidth]{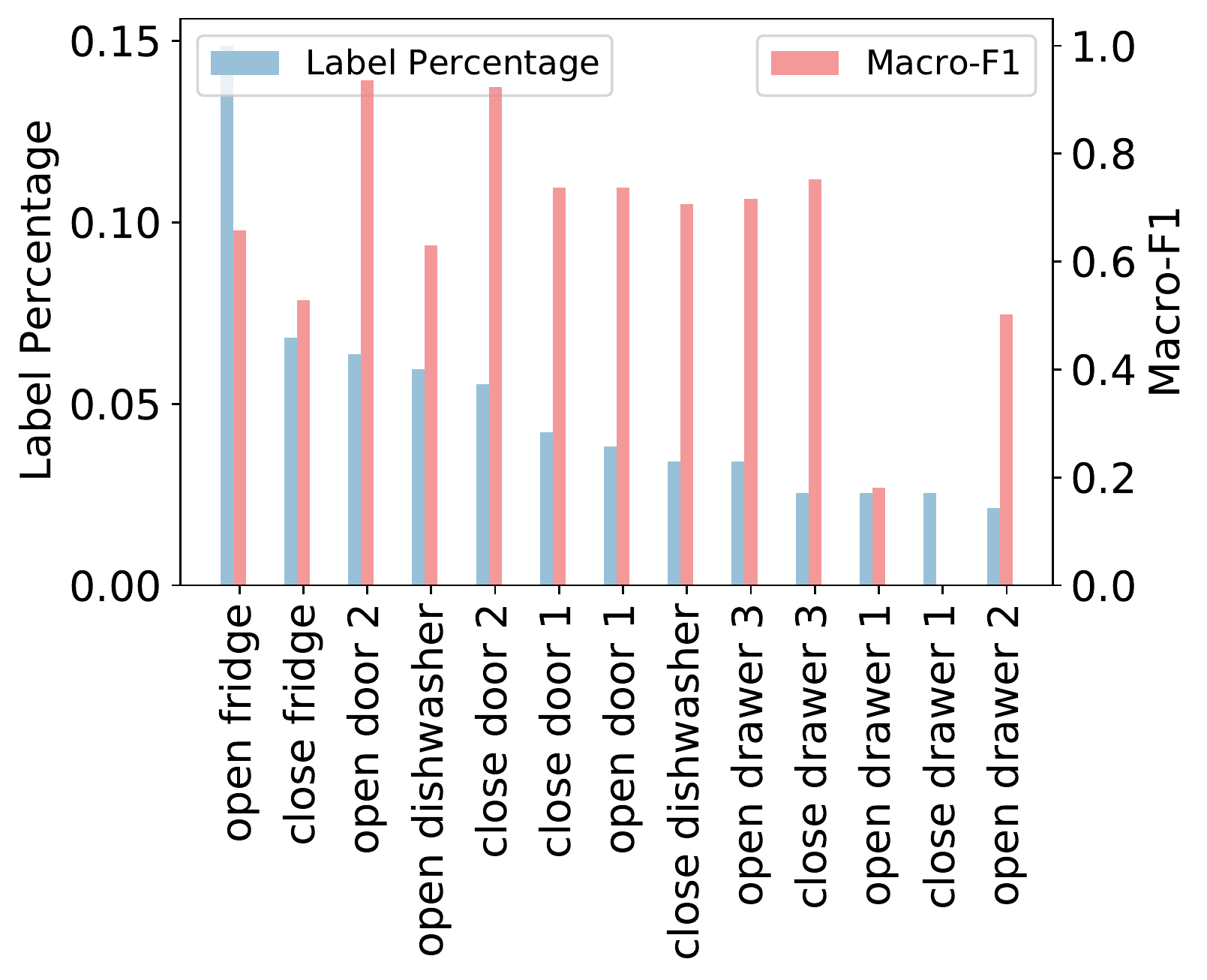}
          \caption{\van}
          \label{fig:van-opp}
      \end{subfigure}
      \begin{subfigure}[b]{0.23\textwidth}
          \centering
          \includegraphics[width=\textwidth]{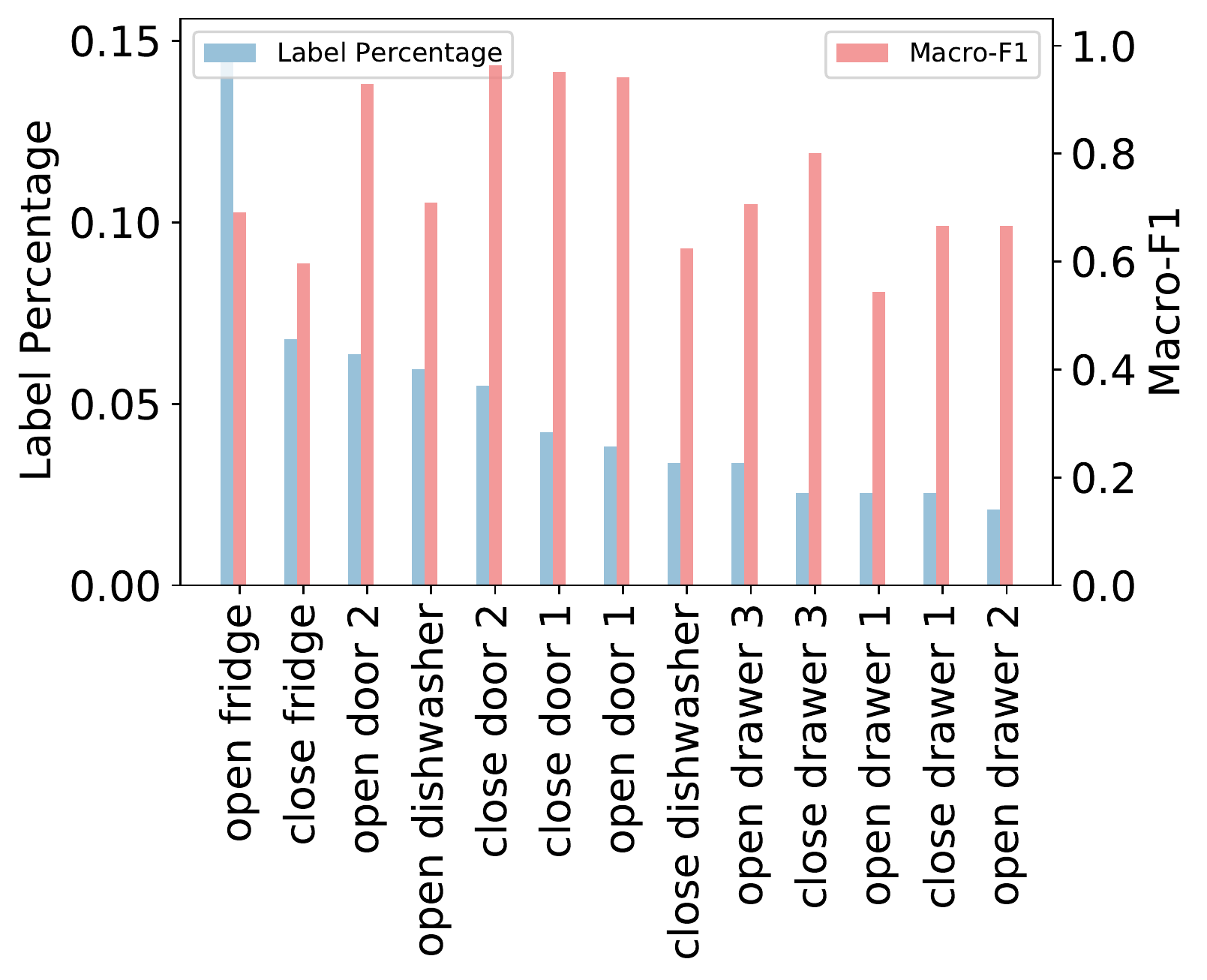}
          \caption{\our}
          \label{fig:ours-opp}
      \end{subfigure}
      \vspace{-3mm}
      \caption{Macro-F1 of example activities with shared label names for \our and \van on Opportunity dataset with long-tail label distribution.}
      \label{fig:long-tail}
\vspace{-5mm}
\end{figure} 

\vspace{-2mm}
\subsection{Model Variants}
We also compare \our with some of its variants to examine the source of the performance gain. For all variants, we use the same encoder for feature extraction as \our.

\begin{itemize}[nosep,leftmargin=*]
\item \noindent \textbf{\van}: We use the same encoder as \our to extract features embedded in the data, and directly append a linear layer for classification without label name modeling.

\item \noindent \textbf{\van + ImageBind embeddings}: We also try directly incorporating ImageBind embeddings into \van. This variant has two separate linear branches at the end. One branch is for classifying the labels, and the other branch predicts embeddings for the label names. During training, apart from the classification cross-entropy loss, we maximize the cosine similarity between the predicted embeddings and the pre-trained ImageBind embeddings. If the label names have multiple words, we use the average ImageBind embedding of each word as the embedding for the entire label name sequence.

\item \noindent \textbf{multi-label classification}: We also try two separate classifiers subsequent to the encoder. The first classifier predicts the original labels, and the second operates as a multi-label classifier that estimates individual tokens within the label sequences. For example, to predict the class ``walk forward'', the second classifier labels ``walk'' and ``forward'' as positive and other tokens as negative. Classification of shared tokens helps learn dependencies across activities, and during testing, we only compare scores from the first classifier for original activity classes. 

\item \noindent \textbf{no aug}: We stay with the label structure decoding architecture but remove all three label augmentations.

\item \noindent \textbf{no token aug}: We stay with the label structure decoding architecture but remove token-level augmentation during training.

\item \noindent \textbf{no embed aug}: We randomly initialize the decoder word embedding layer instead of using ImageBind word embeddings.

\item \noindent \textbf{no seq aug}: The Mhealth dataset~\cite{banos2014mhealthdroid} (publicly available at UCI Machine Learning Repository\footnote{\url{http://archive.ics.uci.edu/ml/datasets/mhealth+dataset}}) rarely has shared tokens in its original label names. We compare the performance of \our on its original non-overlapping label names and pre-trained model-augmented shared label names.
\end{itemize}

As shown in Table~\ref{tab:abl}, we observe significant improvement from only applying a feature encoder to the proposed label structure architecture that decodes label names. Regressing label name embeddings by optimizing a cosine similarity loss with ImageBind embeddings only slightly improves the performance. This demonstrates that 
directly incorporating word embeddings does not explicitly take into account the shared label name structures and loses information when aggregating multiple words into a single label embedding. By contrast, \our generates label sequences which preserves the label structures and encourages knowledge sharing across activities.
Compared with multi-label classification, our label structure decoding approach can preserve the word order (especially for multi-gram) and word correlation in label sequence. 

Moreover, the performances degrade after removing either token-level or embedding-level augmentation (or removing both), which validates their importance in capturing shared word semantics. For sequence-level augmentation, we summarize the original and generated label names from pre-trained model (GPT-4) in Table~\ref{tab:mhealth-label-names}. We compare \our using generated label names against both baselines (Table~\ref{tab:mhealth}) and our model variants (Table~\ref{tab:mhealth-abl}) on the Mhealth dataset. With the help of the automated label generation method, \our demonstrates state-of-the-art performance for HAR datasets without original shared label names. Moreover, we observe that sequence-level augmentation and embedding-level augmentation serve as complementary strategies that synergistically enhance performance.

\vspace{-2mm}
\subsection{Few-Shot Settings}

We further evaluate \our under various few-shot settings. 

\noindent \textbf{Reduced Training Samples.} We randomly reduce the number of samples in the training set from two HAR datasets (Opportunity and UCI-HAR) to 20\%, 40\%, 60\%, and 80\%, and evaluate the macro-F1 on the same original test set. We conducted the experiments for $5$ runs and report both average Macro-F1 as well as standard deviation. Figure~\ref{fig:ds} illustrates the performance trend of \our, \van as well as the best-performing baselines when we vary the size of the training set. As we could observe from the figure, the macro-F1 generally increases as the number of available training samples increases. On top of that, the performance gap between \our and other methods becomes larger when there are fewer training data available, showing that decoding label names helps learn the common structures that are shared across different classes. 

\noindent \textbf{Label Imbalance.} The above experiment reduces training samples for all the classes. Many HAR datasets also naturally have a long-tail distribution where some activities have fewer samples as being more difficult to collect. We also experiment under such label imbalance scenarios as shown in Figure~\ref{fig:long-tail}. We compare \our and the vanilla classification model \van by visualizing example activities with shared tokens. The activity names are sorted in decreasing order by the label percentage in the dataset. The performance grows significantly when adopting the label structure decoding architecture, as decoding label names helps transfer the shared word semantics to those classes with fewer available samples. For example, for the tail classes ``open drawer 1'', ``close drawer 1'', ``open drawer 2'', \van shows a low F1 score (even zero for ``close drawer 1''), while \our substantially improves the performance on these classes, as \our is able to leverage label structures to learn from other classes.

\noindent \textbf{Reduced Window Size.} We also reduce the sampling frequency (window size) on both training and test sets by a factor of 2,4,8 and report the performance of \our, \van as well as the best-performing baselines in Figure~\ref{fig:sam}. 
\our also stays robust with respect to down-sampling factors, as it encourages knowledge transfer via modeling label name structures. 
We observe that our proposed \our consistently outperforms \van and baselines under different down-sampling factors.

\begin{figure}[t]
      \centering
      \begin{subfigure}[b]{0.23\textwidth}
          \centering
          \includegraphics[width=\textwidth]{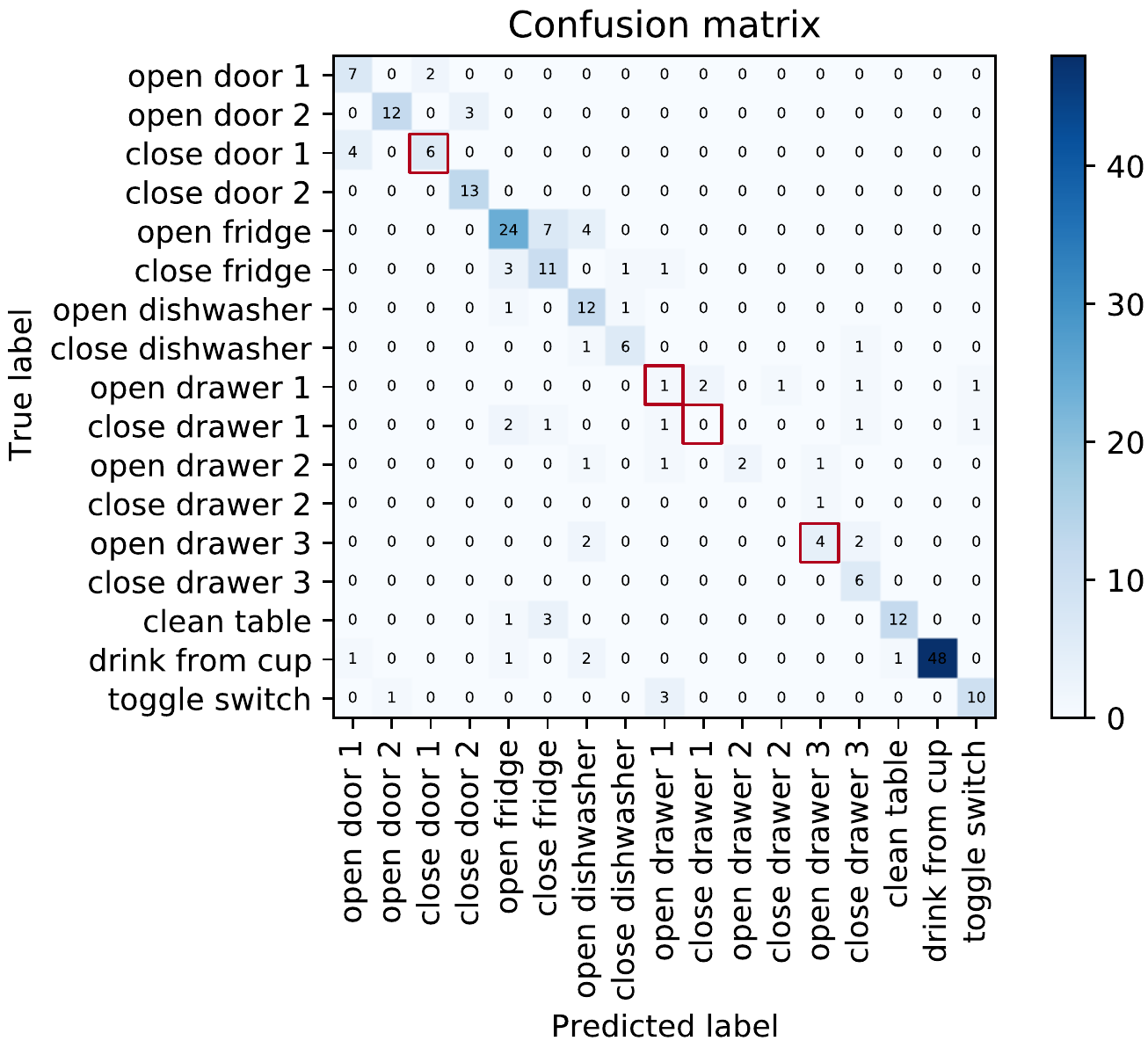}
          \caption{\van}
          \label{fig:van-conf}
      \end{subfigure}
      \begin{subfigure}[b]{0.23\textwidth}
          \centering
          \includegraphics[width=\textwidth]{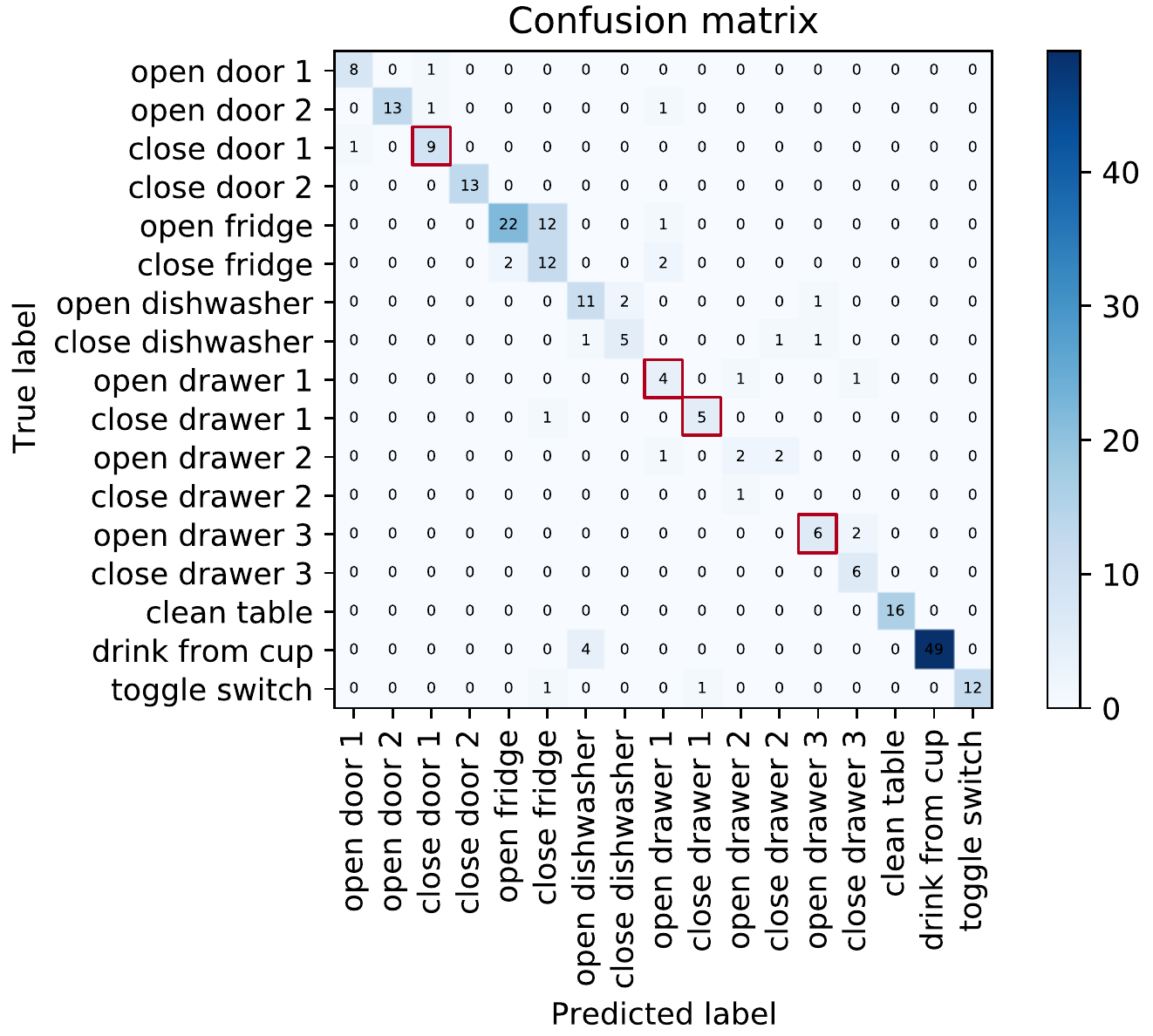}
          \caption{\our}
          \label{fig:ours-conf}
      \end{subfigure}
      \vspace{-3mm}
      \caption{Confusion matrix of \van and \our on Opportunity dataset. \our better discriminates different activities, exemplified by classes with red squares.}
      \label{fig:conf-mat}
\vspace{-5mm}
\end{figure} 

\subsection{Case Study}

In this section, we further explore the benefits of modeling label structures through some case studies. 

\noindent \textbf{Confusion Matrix.} We use Opportunity dataset as an example and show through the confusion matrix in Figure~\ref{fig:conf-mat} that \our better discriminates activities compared with \van, especially for activities with fewer samples. In Figure~\ref{fig:conf-mat}, values at the $i$-th row and $j$-th column represent the number of instances that have ground truth label $i$ and are predicted as label $j$. ``open drawer 1'' instances mispredicted as ``close drawer 1'' are reduced from 2 to 0, and the correctly predicted instances increase from 1 to 4.

\begin{figure}[t]
      \centering
      \begin{subfigure}[b]{0.23\textwidth}
          \centering
          \includegraphics[width=\textwidth]{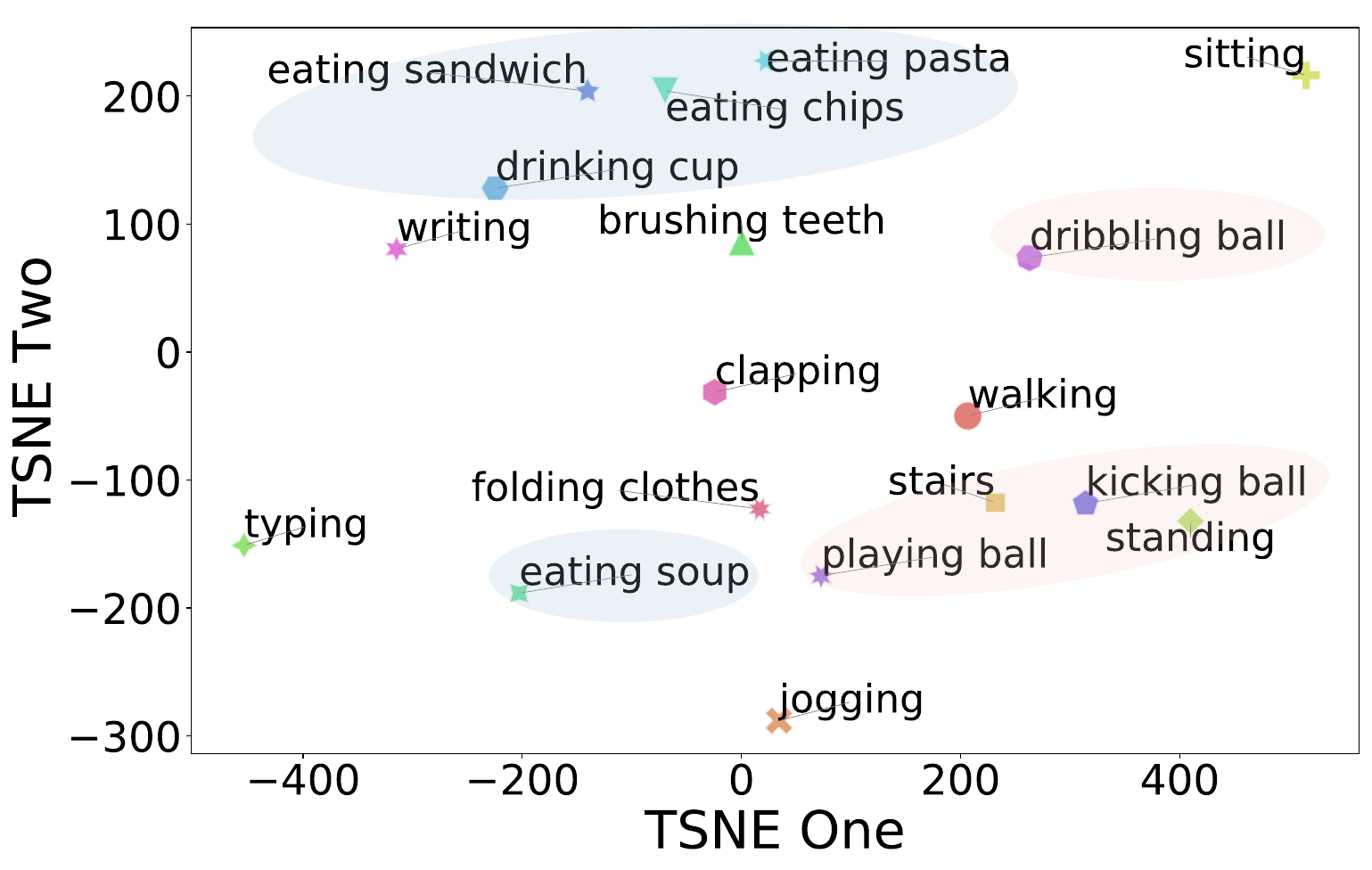}
          \caption{\van}
          \label{fig:van-tsne}
      \end{subfigure}
      \begin{subfigure}[b]{0.23\textwidth}
          \centering
          \includegraphics[width=\textwidth]{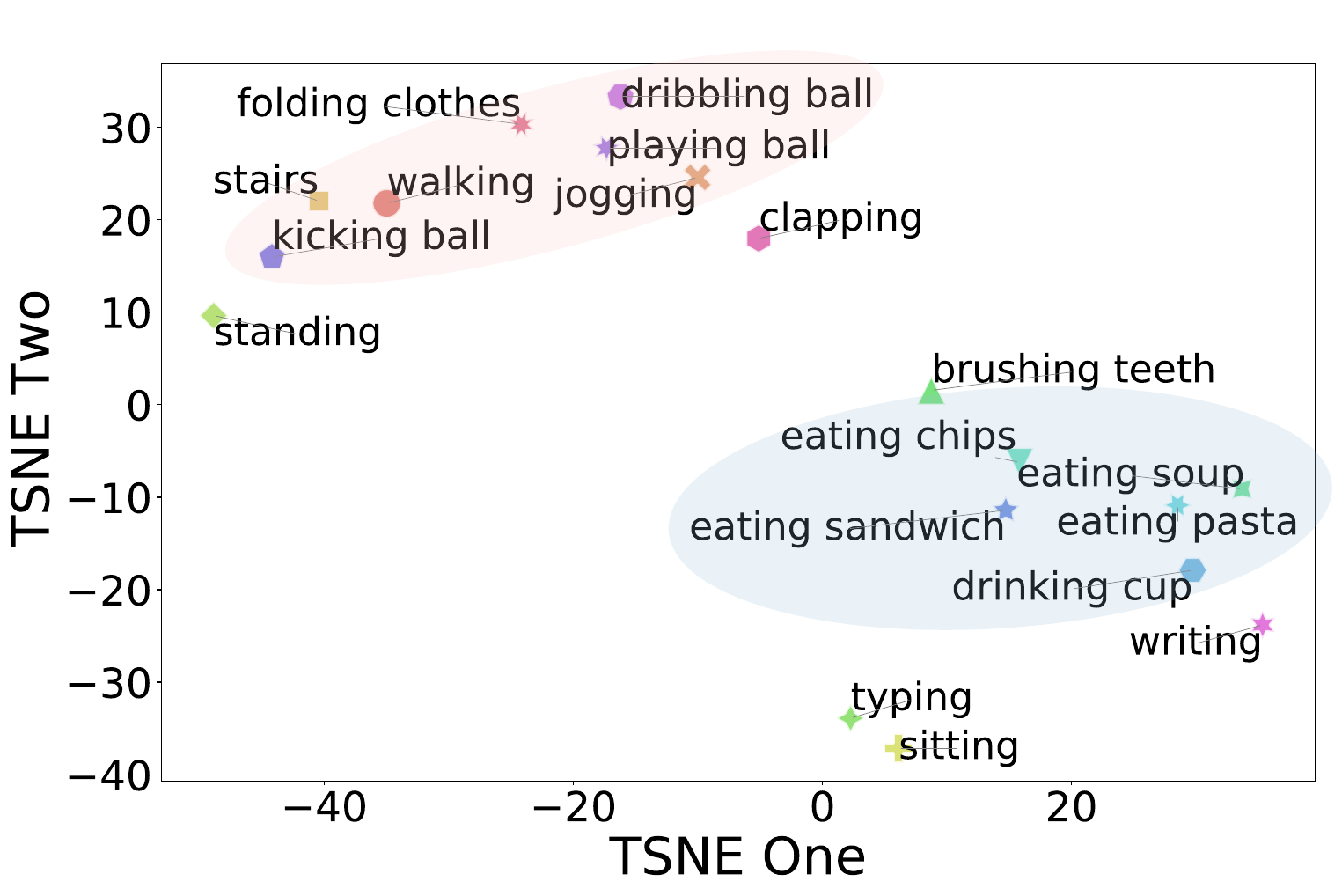}
          \caption{\our}
          \label{fig:ours-tsne}
      \end{subfigure}
      \vspace{-3mm}
      \caption{T-SNE visualization on feature space. \our better preserves the semantics in the feature space.}
      \label{fig:feat-tsne}
\vspace{-3mm}
\end{figure} 

\noindent \textbf{Feature Embedding.} We apply t-SNE visualization to the feature space of \van and \our on the WISDM dataset. We visualize the average feature of each activity, as illustrated in Figure~\ref{fig:feat-tsne}. \van loses the semantic information in the feature space. For example, ``eating soup'' is positioned at a large distance from other ``eating''-related activities. By contrast, \our preserves the label structures in the feature space, indicating a more coherent and precise mapping of related activities.

\vspace{-1mm}
\subsection{Complexity Analysis}
We compare the model complexity of \our and the best-performing deep models TST, TARNet and THAT on PAMAP2 data.
Specifically, we compute the number of parameters, the model size (number of bytes required to store the parameters in the model), and the average running time for a batch of 16 samples (averaged over $10000$ runs). We conducted the complexity analysis on a single NVIDIA RTX A6000 48G GPU. For TST, we only compare the complexity for the supervised fine-tuning phase. As shown in Table~\ref{tab:complexity}, \our has the smallest number of parameters, model size, and average running time, while outperforming more complex deep models. 

\begin{table}[t]
\centering
\small
    \caption{Model complexity analysis.}
    \vspace{-2mm}
	\setlength{\tabcolsep}{1.1mm}{
    \begin{tabular}{r|ccc}
    \hline
        Model & \# of Params & Model Size & Avg Running Time Per Batch  \\
    \hline
        TST & 1.195M & 4.786MB & 0.014s \\
        TARNet & 0.310M & 2.465MB & 0.016s \\
        THAT & 3.207M & 12.828MB & 0.018s \\
        \our & \textbf{0.219M} & \textbf{0.878MB} & \textbf{0.003s} \\

    \hline
    \end{tabular}}
    \label{tab:complexity}
    \vspace{-3mm}
\end{table}
\section{Conclusion}

We proposed a novel HAR approach, \our, that explicitly models the semantic structure of class labels and classifies the activities by decoding label sequence. \our enables knowledge sharing across different activity types via label name modeling and alleviates the challenges of annotated data shortage in HAR, compared with conventional methods that treat labels simply as integer IDs. 
We also design three label augmentation techniques, at token, embedding and sequence levels, to help the model better capture semantic structures across activities.
We evaluated \our on seven HAR benchmark datasets, and the results demonstrate that our model outperforms state-of-the-art methods. 

In the future, we plan to adapt our design to more complex backbone models, as well as image-based or video-based human activity recognition. We also plan to experiment on other types of datasets that also have shared label name structures
, e.g., medical datasets with shared disease names. 
Also, in this work, we assumed that the shared label name structures very likely imply similarity in activity types.
However, the assumption may not hold when we extend the problem scope to simultaneously handling multiple datasets where the same label names may correspond to slightly different data collection settings. 
We believe further investigation to lift such an assumption will offer meaningful insights.

\section{Acknowledgements}
Our work is supported in part by ACE, one of the seven centers in JUMP 2.0, a Semiconductor Research Corporation (SRC) program sponsored by DARPA. Our work is also supported by Qualcomm Innovation Fellowship and is sponsored by NSF CAREER Award 2239440, NIH Bridge2AI Center Program under award 1U54HG012510-01, as well as generous gifts from Google, Adobe, and Teradata. Any opinions, findings, and conclusions or recommendations expressed herein are those of the authors and should not be interpreted as necessarily representing the views, either expressed or implied, of the U.S. Government. The U.S. Government is authorized to reproduce and distribute reprints for government purposes not withstanding any copyright annotation hereon.

\bibliographystyle{ACM-Reference-Format}
\balance
\bibliography{sample-base}

\end{document}